\documentclass[lettersize, journal]{IEEEtran}
\usepackage{amsmath,amsfonts}
\usepackage{algorithmic}
\usepackage{array}
\usepackage[caption=false,font=normalsize,labelfont=sf,textfont=sf]{subfig}
\usepackage{textcomp}
\usepackage{stfloats}
\usepackage{url}
\usepackage{verbatim}
\usepackage{graphicx}
\def\BibTeX{{\rm B\kern-.05em{\sc i\kern-.025em b}\kern-.08em
		T\kern-.1667em\lower.7ex\hbox{E}\kern-.125emX}}
\usepackage{balance}

\usepackage{times}
\usepackage{multicol}
\usepackage[bookmarks=true]{hyperref}
\usepackage{xcolor}
\usepackage{hyperref}
\usepackage{amsmath, amssymb}
\usepackage{amsfonts}
\usepackage{graphicx}
\usepackage{siunitx}
\usepackage{standalone}
\usepackage{booktabs}
\usepackage[ruled,vlined,linesnumbered]{algorithm2e}
\usepackage{mdframed}
\usepackage{fancyvrb}
\usepackage{soul}
\usepackage{dsfont,mathabx}
\usepackage{array, booktabs}
\usepackage{threeparttable}
\usepackage{multirow}

\usepackage{amsthm}
\newtheorem{theorem}{Theorem}

\theoremstyle{definition}

\newtheorem{problem}{Problem}
\newtheorem{definition}{Definition}

\title{Fast and Adaptive Multi-agent Planning under Collaborative Temporal Logic Tasks via Poset Products}

\author{Zesen Liu\textsuperscript{1}, Meng Guo\textsuperscript{1}, Weimin Bao\textsuperscript{2} and Zhongkui Li\textsuperscript{1}
	\thanks{1:The authors are with the State Key Laboratory
		for Turbulence and Complex Systems,
		Department of Mechanics and Engineering Science,
		College of Engineering, Peking University, Beijing 100871, China.
		E-mail: \texttt{1901111653, meng.guo, zhongkli@pku.edu.cn}}
		\thanks{2:The auther is with Science and Technology Commission of China Aerospace Science and Technology Corporation, Beijing 100048, China.
			E-mail:\texttt{BAOWEIMIN@CASHQ.AC.CN}
	}
}
\begin{document}

\date{}

 
\maketitle

\begin{abstract}
	
	Efficient coordination and planning is essential for large-scale multi-agent systems
	that collaborate in a shared dynamic environment.
	Heuristic search methods or learning-based approaches often
	lack the guarantee on correctness and performance.
	Moreover, when the collaborative tasks contain both spatial and temporal requirements,
	e.g., as Linear Temporal Logic (LTL) formulas,
	formal methods provide a verifiable framework for task planning.
	However, since the planning complexity grows exponentially with the number of
	agents and the length of the task formula,
	existing studies are mostly limited to small artificial cases.
	To address this issue,
	a new planning paradigm is proposed {in this work}
	for system-wide temporal task formulas that are released online and continually.
	It avoids two common bottlenecks in the traditional methods, i.e.,
	(i) the direct translation of the complete task formula to the
	associated B\"uchi automaton;
	and (ii) the synchronized product between the B\"uchi automaton
	and the transition models of all agents.
	Instead, an adaptive planning algorithm is proposed that
	computes the product of relaxed partially-ordered sets (R-posets) on-the-fly,
	and assigns these subtasks to the agents subject to the ordering constraints.
	It is shown that the first valid plan can be derived
	with a polynomial time and memory complexity w.r.t. the system size and the formula length.
	Our method can take into account task formulas with a length of more than~$400$
	and a fleet with more than $400$ agents,
	while most existing methods fail at the formula length of~$25$ within a reasonable duration.
	The proposed method is validated on large fleets of service robots in both
	simulation and hardware experiments.
\end{abstract}


\section{Introduction} 

Recent advances in computation, perception and communication
allow the deployment of autonomous robots in large, remote and hazardous environments,
e.g., to assist service staff in hospitals~\cite{jemal2015multi},
to maintain offshore drilling platforms~\cite{cliff2015online},
to monitor and assist construction sites~\cite{zhang2009collaborative}.
Furthermore, fleets of heterogeneous robots,
such as unmanned ground vehicles and unmanned aerial vehicles,
are deployed to accomplish tasks that are otherwise too inefficient
or even infeasible for a single robot~\cite{arai2002advances}.
Not only the overall efficiency of the team can be significantly improved
by allowing the robots to move and act concurrently~\cite{toth2002overview};
but also the capabilities of the team can be greatly extended by
enabling multiple robots to directly collaborate on a task~\cite{fink2008multi}.
Recent works have demonstrated such potentials preliminary for simple
tasks such as collaborative exploration~\cite{doi:10.1126/scirobotics.ade9548}, formation control~\cite{varava2017herding},
object transportation~\cite{doi:10.1126/scirobotics.abf1628} and pursuer-evader games \cite{doi:10.34133/research.0246}.
The task planning problem for multi-robot systems is in general NP-hard~\cite{1997Task},
due to the inherent combinatorial nature of robot-task assignment and various constraints such as capabilities and deadlines.
The standard approach is to formulate a mixed integer linear programs (MILP)
over the integer variables and constraints~\cite{agrawal2016scalable}.
Whereas being sound and optimal, these methods are applicable to only small-scale systems.
Thus, extensive work can be found on designing meta-heuristic algorithms for finding
sufficiently good solutions in a reasonable time,
e.g., genetic algorithms~\cite{KESHANCHI20171}, colony optimization~\cite{li2017multi,doi:10.34133/2020/1762107},
particle swarm optimization~\cite{yan2021task,8280872},
learning-based algorithms~\cite{wells2019learning,pmlr-v70-omidshafiei17a,liu2019task},
or large language models~\cite{zhang2023building,ruan2023tptu}.
However, these methods often lack a formal guarantee on the correctness and quality of the planning
results.

Moreover, to specify more complex tasks,
many recent works propose to use formal languages
such as Linear Temporal Logic (LTL)~\cite{Katoen2008Principles}, Computation Tree Logic (CTL)~\cite{koymans1990specifying},
and Signal Temporal Logic (STL)~\cite{maler2004monitoring}, as an intuitive yet
powerful way to describe both spatial and temporal requirements on
global~\cite{luo2021abstraction} or local~\cite{guo2016task} behaviors.
Notably, the works in~\cite{luo2021temporal, sahin2019multirobot, jones2019scratchs}
formulate MILP by a central planning unit given different system models and task constraints;
the works in~\cite{2016Decomposition,schillinger2018simultaneous,kantaros2020stylus,YU2022105130,li2023fast}
instead propose various search algorithms over the state or solution space of the whole system.
However, the aforementioned planning methods are often executed {offline}
for a set of predefined {static} tasks.
A particularly challenging scenario is when the system operates indefinitely,
i.e., new tasks are released or canceled \emph{dynamically} and \emph{continually}
by external demand \cite{7306591};
or certain target features related to the tasks can change location
during run time~\cite{doi:10.34133/research.0246}.
This would require the fleet to adaptively change their task plans online to modify
existing assignments and incorporate new tasks.
Thus, the aforementioned methods become inadequate as the sequence of tasks is infinite
and their specifications are unknown beforehand.
Recursive application of the centralized methods in a naive way leads to
not only intractable computation complexity,
but also inconsistent or even oscillatory assignments.
Thus, an efficient and adaptive planning scheme is essential for multi-robot systems
that collaborate in a dynamic environment \cite{8901075,10102342,choudhury2022dynamic}
or an unknown environment \cite{9884983}.

\subsection{Related Work} \label{related-work}
The standard framework for planning under temporal tasks is based on the model-checking algorithm~\cite{Katoen2008Principles}:
First, the task formulas are converted to a Deterministic Robin Automaton (DRA)
or Nondeterministic B\"uchi Automaton (NBA),
via off-the-shelf tools such as~\texttt{SPIN}~\cite{ben2010primer} and~\texttt{LTL2BA}~\cite{10.1007/3-540-44585-4_6}.
Second, a product automaton is created between the automaton of formula and the models of all agents,
such as weighted finite transition systems (wFTS)~\cite{Katoen2008Principles}, Markov decision processes~\cite{6702421}
or Petri nets~\cite{7294096}.
Last, certain graph search or optimization procedures are applied to find a valid and accepting plan
within the product automaton, such as nested-Dijkstra search~\cite{schillinger2018simultaneous}, integer programs~\cite{9663414},
auction~\cite{schillinger2019hierarchical} or sampling-based~\cite{kantaros2020stylus,kantaros2018sampling}.

Thus, the fundamental step of all aforementioned methods is to translate the task formula into
the associated automaton.
This translation may lead a double-exponential size
w.r.t. the formula length as shown in~\cite{10.1007/3-540-44585-4_6}.
The only exceptions are GR(1) formulas~\cite{10.1007/11609773_24},
of which the associated automaton
can be derived in polynomial time but only for limited cases.
In fact, for general LTL formulas with length more than~$25$,
it takes more than~$2$ hours and $13 GB$ memory to compute the associated NBA via~\texttt{LTL2BA}.
Although recent methods have greatly reduced the planning complexity in other aspects,
the length of considered task specifications remains limited due to this translation process.
For instance, the sampling-based method~\cite{kantaros2020stylus,kantaros2018sampling}
	avoids creating the complete product automaton via RRT sampling,
	of which the largest simulated case has $400$ agents and
	the task formula has a maximum length of $14$.
The planning method~\cite{schillinger2018simultaneous} decomposes the resulting task automaton
into independent sub-tasks for parallel execution.
The simulated case scales up to~$100$ robots and a task formula of maximum length~$18$.
Moreover, other existing works such
as~\cite{2021Reactive,2018Multi,6942756,7487762} mostly
{consider} task formulas of length around~$6$-$10$.
This limitation hinders its application to more complex robotic missions.


This drawback becomes even more apparent for dynamic scenes,
where contingent tasks specified as LTL formulas are triggered by external observations
and released to the whole team online.
In such cases, most existing approaches compute
the automaton associated with the new task, of which the synchronized
{product} with the current automaton is derived, see e.g.,~\cite{6942756, 7487762}.
Thus, the size of the resulting automaton equals to the product
of \emph{all} automata associated with the contingent tasks,
which is clearly a combinatorial blow-up.
Consequently, the amount of contingent tasks that can be
handled by the aforementioned
approaches is limited to hand-picked examples.

\subsection{Our Contribution}
To overcome this curse of dimensionality in the size of tasks and agents,
we propose a new paradigm that is \emph{fundamentally} different from
the model-checking-based methods.
First, for a syntactically co-safe LTL (sc-LTL) formula that is a conjunction of numerous sub-formulas,
we calculate the R-posets of each sub-formula as a set of subtasks and
their partial temporal constraints.
Then, an efficient algorithm is proposed to compute the product of
R-posets associated with each sub-formula.
{The resulting product of R-posets is complete in the sense}
that it retains all subtasks from each R-poset along with their partial orderings
and resolves potential conflicts.
Given this product,
a task assignment algorithm called the time-bound contract net (TBCN) is proposed to
assign collaborative subtasks to the agents,
subject to the partial ordering constraints.
Last but not least, the same algorithm is applied online to dynamic scenes
where contingent tasks are
triggered and released by online observations.
It is shown formally that the proposed method has a polynomial time
and memory complexity to derive the first valid plan
w.r.t. the system size and formula length.
Extensive large-scale simulations and experiments are conducted
for a hospital environment where service robots react
to online requests such as collaborative transportation,
cleaning and monitoring.

Main contribution of this work is three-fold:
(i) a systematic method is proposed to tackle task formulas with length more than~$400$,
which overcomes the limitation of existing translation tools that can only process
formulas of length less than~$25$ in reasonable time;
(ii) an efficient algorithm is proposed to decompose and integrate
contingent tasks that are released online,
which not only avoids a complete re-computation of the task automaton
but also ensures a polynomial complexity to derive the first valid plan;
(iii) the proposed task assignment algorithm is fully compatible with
both static and dynamic scenarios with interactive objects.

\subsection{Problem Statement}
{Consider a multi-agent system with heterogeneous capabilities,
	a series of sc-LTL task formulas that are released online,
	and a set of interactive objects in the dynamic environment.
	The objective is to generate a task plan for the system online such that
	these tasks are satisfied with high efficiency.}

{For instance as shown in Sec.~\ref{simulation},
	a fleet of heterogeneous service robots is deployed in the hospital environment.
	Different tasks such as transportation of goods or patients, cleaning and maintenance
	are released online continuously.
	Many of such tasks contain numerous subtasks  with ordering constraints
	that require direct collaboration of different robots.
	An efficient coordination algorithm is proposed such that these subtasks
	are assigned and fulfilled online in a timely manner.
}

\section{Results}

{In this section, we present the proposed solution briefly, where we first give
	a definition of task specification and introduce the core method of computing the R-poset product.
	Then, an efficient assignment algorithm is introduced under these posets with temporal and spatial constraints.
	The simulation and hardware experiment results are presented,
	against several strong baselines for different sizes of fleets and task complexities.
	Technical details and derivations can be found in the section of ``Materials and Methods''.}

\subsection{Task specification and R-poset product}\label{Result_taskspecification}

{We briefly introduce the syntax of Linear Temproal Logic (LTL)
	used for task specifications.}
The basic ingredients of LTL formulas are a set of atomic propositions $AP$
in addition to several Boolean and temporal operators.
Atomic propositions are Boolean variables that can be either true or false.
The syntax of LTL is defined as:
$\varphi \triangleq \top \;|\; p \;|\; \varphi_1 \vee \varphi_2 \;|\; \varphi_1 \wedge \varphi_2  \;|\; \neg \varphi  \;|\; \bigcirc \varphi  \;|\;  \varphi_1 \,\textsf{U}\, \varphi_2\,|\,\Box \varphi\,|\, \Diamond\varphi$
where $\top\triangleq \texttt{True}$; $p \in AP$ is the alphabet;
$\vee$ (\emph{conjunction}), $\wedge$ (\emph{disjunction}), and  $\neg$ (\emph{negation}) are the logical operators;
$\Diamond$ (\emph{eventually}), $\bigcirc$ (\emph{next}),
$\textsf{U}$ (\emph{until}) are the temporal operators;
$\Box$ (\emph{always}),  $\Rightarrow$ (\emph{implication}) are the derived operators.
A special class of LTL formula called \emph{syntactically co-safe} formulas (sc-LTL)~\cite{Katoen2008Principles, belta2017formal} only contain the temporal operators $\bigcirc$, $\textsf{U}$ and $\Diamond$.
A complete description of the semantics and syntax of LTL can be found in~\cite{Katoen2008Principles}.

\begin{figure*}[ht]
	\centering
	\includegraphics[width=1\linewidth]{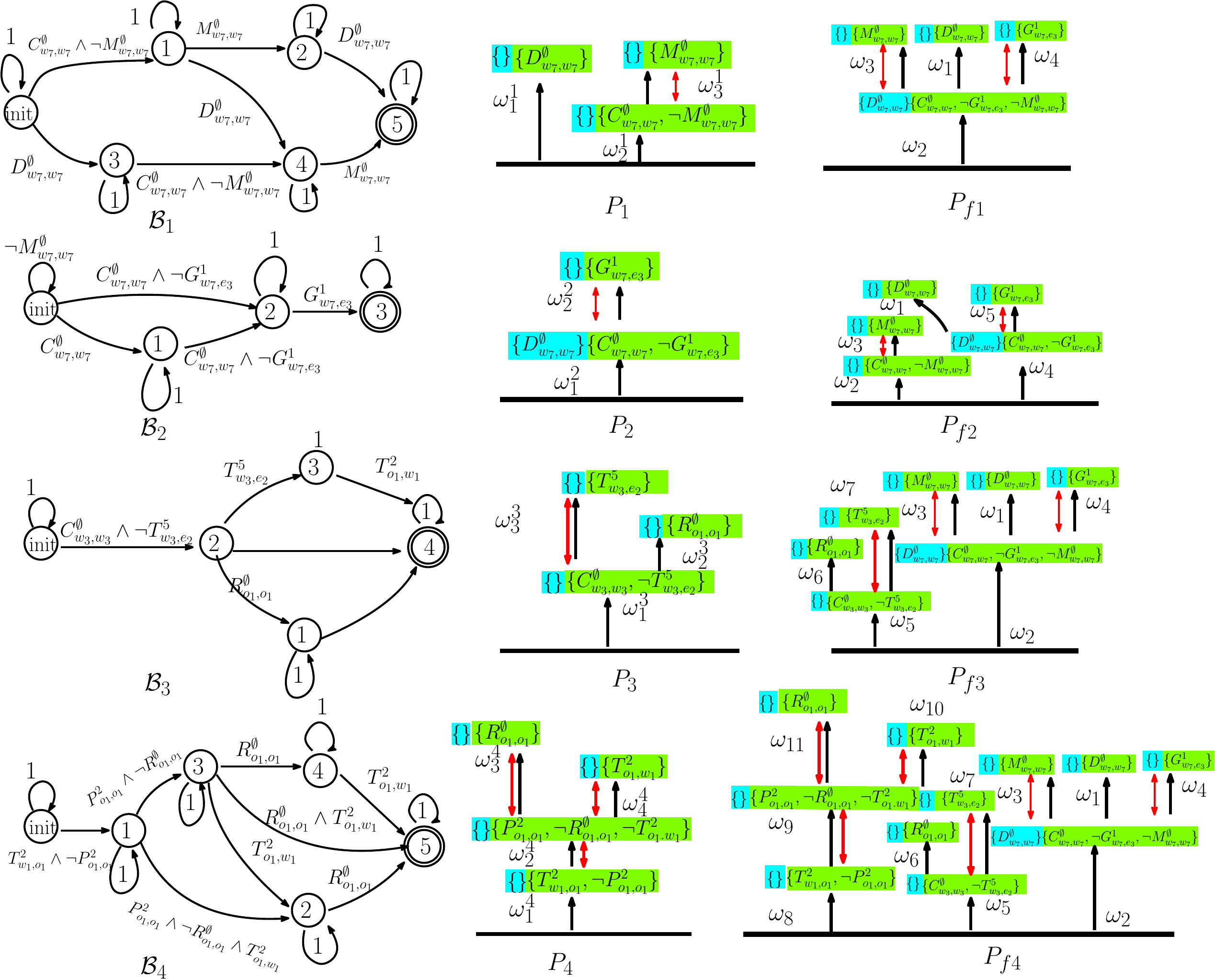}
	\caption{Example of \emph{Poset-Prod} associated with four
		formulas $\varphi_{1}=  \Diamond D^{\emptyset}_{w7,w7}  \land \Diamond(C^\emptyset_{w7,w7}
		\land \neg M^{\emptyset}_{w7,w7}
		\land \Diamond M^{\emptyset}_{w7,w7} )$,
		$\varphi_{2}=\Diamond (C^\emptyset_{w7,w7} \land \neg G^1_{w7,e3} \land
		\Diamond G^1_{w7, e3}) \land \neg D^\emptyset_{w7,w7} \textsf{U} C^\emptyset_{w7,w7} $
		, $\varphi_3=\Diamond( C^\emptyset_{w3,w3}\land\neg T^5_{w3,e2}\land\Diamond D^\emptyset_{w3,w3}\land\Diamond T^5_{w3,e2})$
		and $\varphi_{4}=\Diamond(T^{2}_{w1,o1}\land
		\neg P^{2}_{o1,o1} \land \Diamond (P^{2}_{o1,o1} \land \neg R^{\emptyset}_{o1,o1} \land
		\Diamond T^{2}_{o1,w1} \land \Diamond R^{\emptyset}_{o1,o1}))$
		\textbf{Left:} $\mathcal{B}_1, \cdots,\mathcal{B}_4$ are the NBAs of formula $\varphi_1, \cdots,\varphi_4$;
		\textbf{Middle:} $P_1,\cdots,P_4$ are the R-posets of $\mathcal{B}_1,\cdots,\mathcal{B}_4$;
		\textbf{Right:} $P_1\otimes P_2=\{P_{f1}, P_{f2}\}$, $P_{f_3}\in P_{f_1}\otimes P_3$ and $P_{f_4}\in P_{f_3}\otimes P_{4}$.
	}
	\label{fig:poset_product}
\end{figure*}

{Given a series of sc-LTL formulas that are released online, most existing methods would
	combine the formulas with $\wedge$ (\emph{conjunction}) and convert them into
	a Nondeterministic B\"{u}chi Automaton (NBA).
	It results in a graph structure that consists of states, transitions, guards, inital states, and final states.
	However, since its size is double exponential to the length of formula~$\varphi$ as proven in \cite{10.1007/3-540-44585-4_6},
	it quickly becomes intractable with more task formulas.
	Instead, we compute the product of R-posets associated with these formulas,
	called the \emph{Poset-prod} (denoted by~$\otimes$).
	The R-poset $P=(\Omega,\preceq,\neq)$ is a high-level abstraction of NBA,
	proposed in our earlier work~\cite{LIU2024111377},
	which consists of a set of subtasks $\Omega$ and their partial
	relations as less equal $\preceq$ and conflict $\neq$.
	It has been proven therein that if the subtasks are executed under the partial relations,
	the resulting traces satisfy the NBA.
	Once a new task formula is released, it is transformed into a new R-poset $P_2$ and
	its product with the current R-poset $P_1$ is computed.
	More specifically, given $P_1,P_2$, the {Poset-prod}
	returns a new set of R-posets~$\mathcal{P}=P_1\otimes P_2$ that satisfies both $P_1$ and $P_2$,
	which is computed by iterating the following two procedures:
	\textbf{Task Composition} and \textbf{Relation Update}.
	The first procedure is to create a group of subtasks
	as a combination between the subtasks in $P_2$ and the subtasks of~$P_1$
	that have not been executed.
	A depth-first-search algorithm is proposed to
	gradually add the subtasks
	along with the corresponding mapping function.
	The second procedure is to calculate the partial relations between the subtasks
	such that the ordering constraints among the subtasks
	in the composed product are consistent without conflicts.
	Consequently, the overall poset is given by~$P_{final}=(\Omega_{f},\preceq_{f},\neq_{f})$,
	which is iteratively computed each time a new poset is added.}

{The proposed \emph{Poset-prod} method outperforms the tradition method
	both in computational efficiency and performance.
	This improvement becomes even more pronounced when the number and length of sub-formulas increase.}
This is because the time of generating NBAs of $\varphi_1,\varphi_2,\cdots$ grows linearly, while
the time of converting $\varphi_1\land\varphi_2\land\cdots $ into NBA grows exponentially.
Moreover, for a new added formula, the algorithm updates the final R-poset based on the previous result,
which ensures the performance for online cases.
Finally, it is an anytime algorithm that the algorithm can calculate an R-poset
within linear complexity for multiple sub-formulas at the expense of optimality.
The concrete complexity analysis of \emph{Poset-pord}
is shown in Sec.~\ref{discussion}, and the definition
of R-poset and label is shown in Sec.~\ref{sec:method}.

To give an example,
	consider four sub-formulas $\varphi_1,\varphi_2,\varphi_3,\varphi_4$
	as shown in Fig.~\ref{fig:poset_product}.
	Their NBAs $\mathcal{B}_1,\cdots,\mathcal{B}_4$ can be transformed into four R-posets $P_1,\cdots,P_4$.
	The first round of \emph{Poset-prod} is between $P_1$ and $P_2$ as $P_1\otimes P_2=\{P_{f1}, P_{f2}\}$.
	Then, the following rounds of \emph{Poset-prod} are performed between the results of the previous round and
	the next R-poset as $P_{f1}\otimes P_3$,
	of which the first solution is denoted by~$P_{f3}$.
	At the expense of some optimality, the algorithm can go to next round
	before all results of $P_{f2}\otimes P_3$ are generated, as it is an anytime-algorithm. 
	Finally, we can derive $P_{f4}$ by computing $P_{f3}\otimes P_4$ as the final R-poset which
	satisfies $P_1,\cdots, P_4$.
	{Note that the time to generate the NBA associated with $\varphi_1$, $\bigwedge^2_{i=1}\varphi_i$,
		$\bigwedge^3_{i=1}\varphi_i$ and $\bigwedge^4_{i=1}\varphi_i$ grows significantly
		with a duration of $0.058s, 0.076s,  1.56s,25.56s$ via \texttt{LTL2BA}~\cite{10.1007/3-540-44585-4_6}.
		By contrast, the time to compute these products
		$P_1\otimes P_2, P_1\otimes P_2\otimes P_3, P_1\otimes P_2\otimes P_3 \otimes P_4$
		grows slower with a duration of $0.199s,0.279s, 0.385s$.
		Thus, our method can deal with a much larger number of tasks online.
		Detailed comparisons between our method and traditional methods can be found in Sec.\ref{scale_analysis}.

	\subsection{Online Subtask Assignment under Complexity Constraints}
	
	\begin{figure*}
		\centering
		\includegraphics[width=0.8\linewidth]{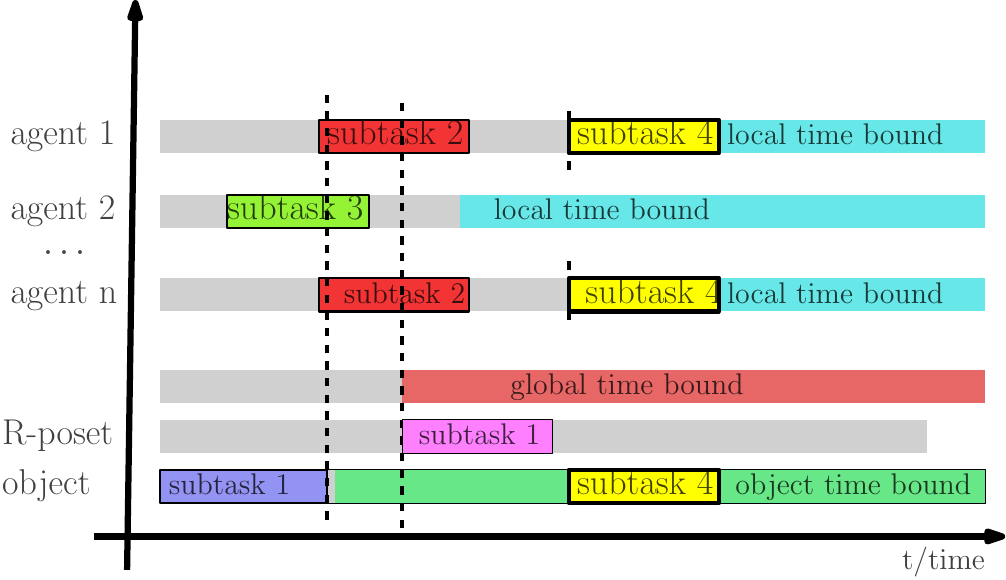}
		\caption{The bidding process of assigning subtask 4 under the global time
			bound (when the partial relations in R-poset are satisfied),
			the time bound for object (when the object is reachable)
			and local time bounds (when the agents get ready).}
		\label{fig:task_assignment}
		\vspace{-0.1in}
	\end{figure*}
	
	Once the set of R-posets that satisfies all sub-formulas is generated, 
	a series of subtasks in the R-poset should be assigned to the agents under several complexity constraints,
	including temporal constraints from R-poset, objects constraints and cooperative constraints.
	Similar to the classical Contract Net method \cite{1675516},
	we propose an efficient and sub-optimal assignment algorithm called Time Bound Contract Net 
	(TBCN). The main difference is that all these constraints are represented uniformly
	by time bounds in our methods, and an example is shown in Fig.~\ref{fig:task_assignment}~.
	The final assignment is a group of timed sequence of robot actions $\mathcal{J}=[J_1,\cdots,J_n]$,
	where $J_n=[(t_k,\omega_{k},a_k),\cdots]$ indicates that agent~$n$ will execute action~$a_k$
	at time $t_k$ to satisfy subtask $\omega_k$, 
	such that the newly-released tasks are fulfilled.
 
 	TBCN consists of three steps:
	\textbf{Initialization}, \textbf{Computation of Feasible Subtasks}
	and \textbf{Online Bidding}. 
	As the partial constraints of final R-poset $P_{final}$ might be changed
	after computing the product,
	the subtasks that are conflicting with the updated partial orders are removed   
	in \textbf{Initialization}.
	Then, the last two steps are iterated: 
	the set of subtasks whose partial orders are satisfied given the current assignment $\mathcal{J}$ in \textbf{Computation of Feasible Subtasks};
	Consequently, a linear program is formulated and solved in the \textbf{Online Bidding}, 
	to choose one subtask from the feasible set.
	Thus, the resulting execution time and action plans are added to $\mathcal{J}$.

	\subsection{Numerical Simulation}
	\label{simulation}
	\begin{figure*}
		\centering
		\includegraphics[width=1\linewidth]{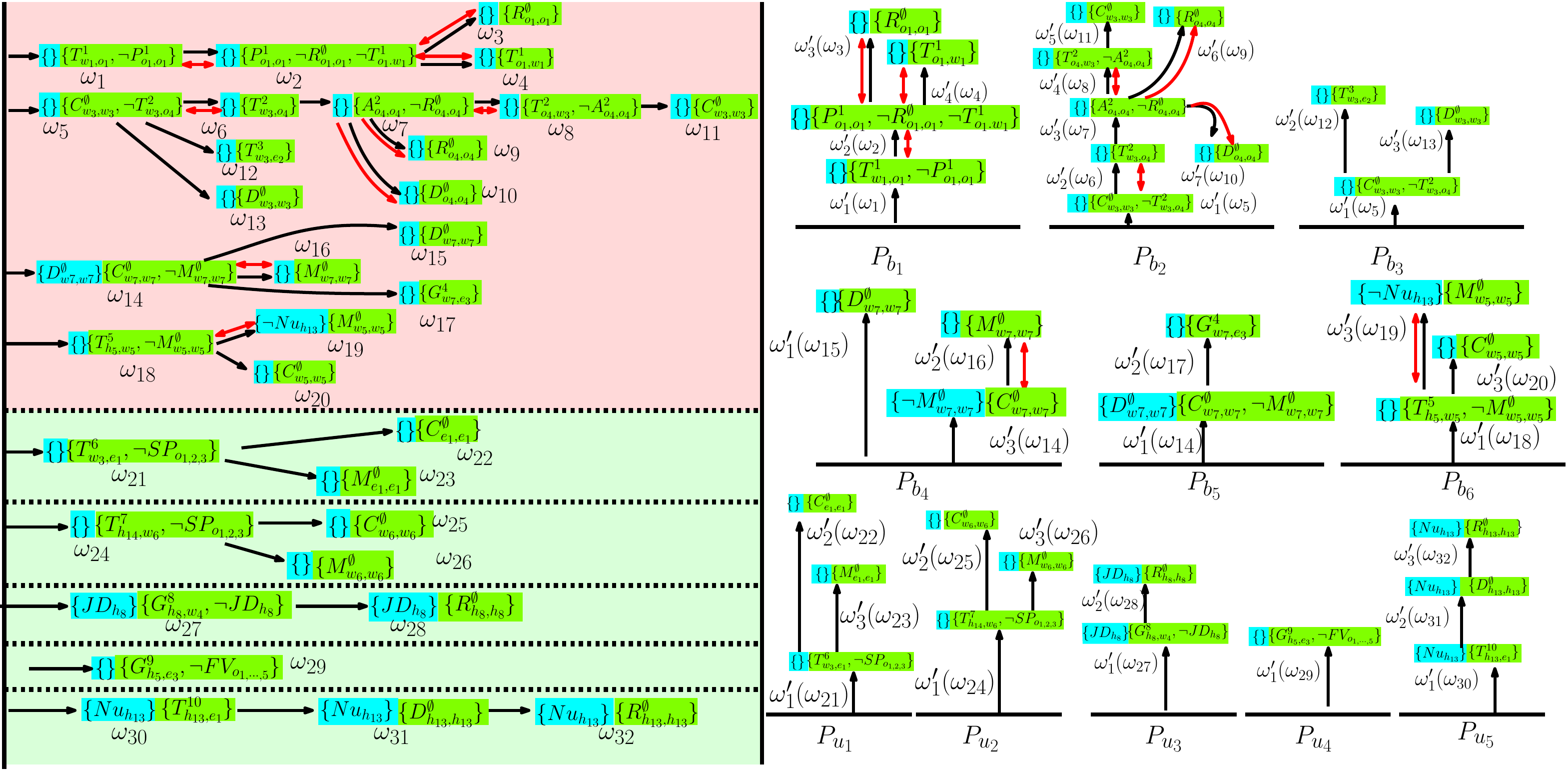}
		\caption{Illustration of computing the product posets.
			\textbf{Left}: The final R-poset
			computed from the initial sub-formulas and online sub-formulas (dashed lines);
			\textbf{Right}: $P_{b_1},\cdots,P_{b_6}$ are the R-posets associated
			with the initial sub-formulas and $P_{u_1},\cdots,P_{u_5}$ are associated with the online sub-formulas.}
		\label{fig:simulation_poset}
		\vspace{-0.1in}
	\end{figure*}
	
	As shown in Fig.~\ref{fig:trajectory_gantt},
		the hospital environment consists of the wards, the operating rooms, the hall,
		the exits and the hallways. The multi-agent system is employed with 3 Junior Doctors, 6 Senior
		Doctors and 8 Nurse, and 4 types of patients are treated as interactive objects including Family
		Visitors, Vomiting Patients, Senior Patients and Junior Patients. The detailed mappings between
		agents, action, objects and their labels are shown in Table~\ref{table:definition_function}.
	Furthermore, various types of tasks are considered, such as ``Go the rounds of the wards and provide medicine",
	``Check and record the patient status", and ``Prepare and perform an operation on a patient".
	Some tasks are released online under certain conditions.
	For instance, when a patient vomits at a region, the task
	"Take the patient into ward and check his status. Meanwhile, the doctors should
	not enter this region until it has been cleaned".
	The sc-LTL formula associated with the complete task is given by $\varphi=\varphi_{b1}\land\cdots\varphi_{b_6}\land\varphi_{VP}\land\varphi_{JP}\land\varphi^1_{SP}\land\varphi^2_{SP}\land\varphi_{FV}$, which has a total length of 62.
	Detailed description of formulas are shown in Table~\ref{table:definition_formulas}.

	\begin{figure*}[t]
		\centering
		\includegraphics[width=0.9\linewidth]{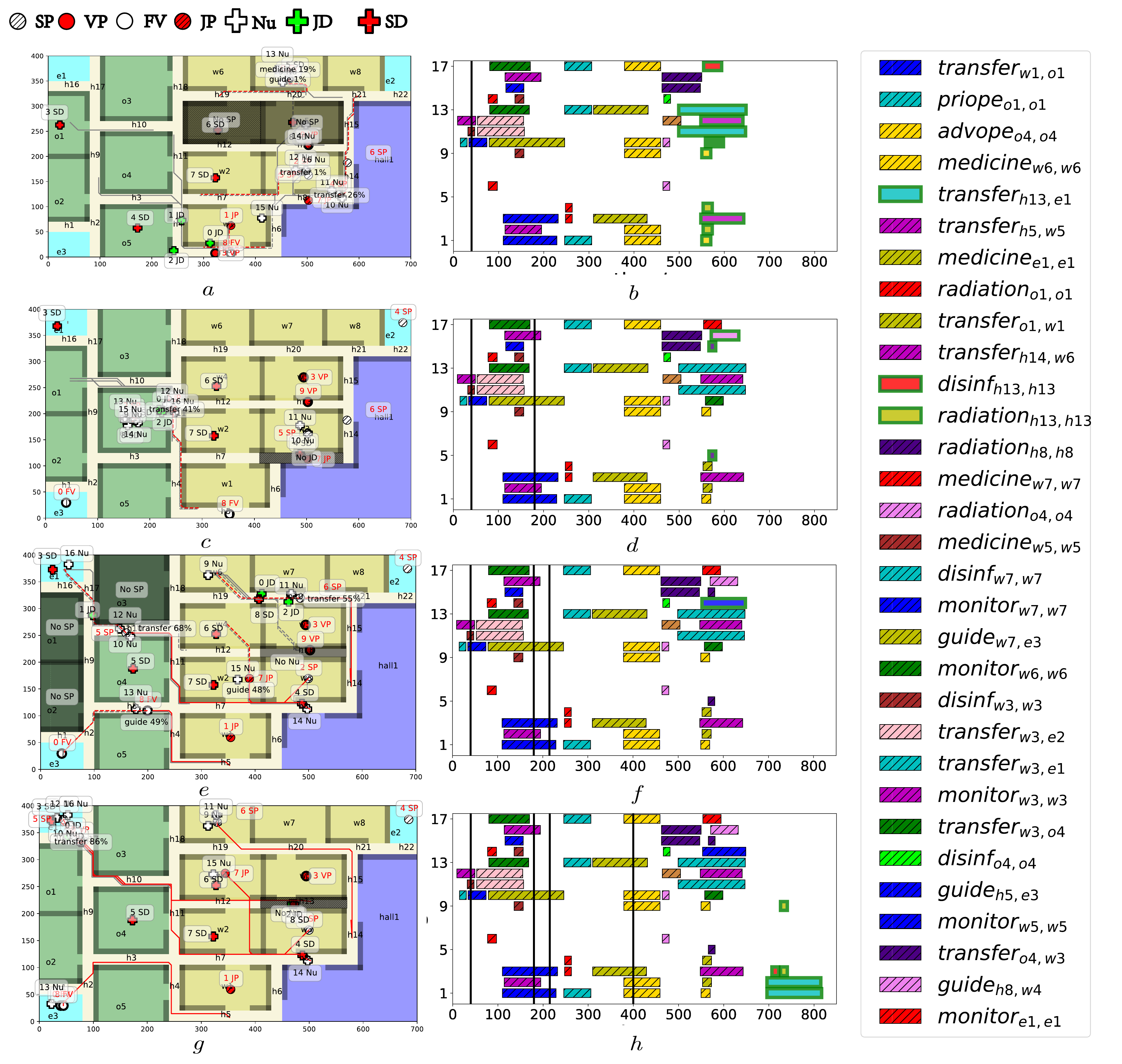}
		\caption{
			\textbf{Left}: Snapshot of agent trajectories at $40,180,215,400s$ when new tasks are added online.
			\textbf{Right}: The Gantt graph of task assignment at these time instants,
			as highlighted in green boxes.}
		\label{fig:trajectory_gantt}
		\vspace{-0.1in}
	\end{figure*}
	
	As shown in Fig.~\ref{fig:simulation_poset}, each subtask $\omega_i$
	consists of the constraints before execution (in blue)
	and during execution (in green).
	The directed black arrows denote the ordering constraints,
	while the bidirectional red arrows for the conflicting constraints.
		The mapping  from the subtasks within the input R-posets
		to the subtasks within the final R-poset is shown in brackets as~$\omega'_1(\omega'_1):\omega'_1\rightarrow\omega_1$.
	Note that most of R-posets on the right are directly incorporated
	into the final R-posets on the left, such as $P_{b_1}, \cdots ,P_{b_6}, P_{u_1},\cdots,P_{u_5}$.
	{This is due to the fact that their subtasks are independent without ordering constraints,
		which allows for parallel execution}. Additionally, some subtasks on the left represent
	the same subtasks on the right. For example, $\omega'_1$ of $P_{b_2}$ and
	$\omega'_1$ of $P_{b_3}$ representing both $\omega_5$,
	since their ordering constraints (in green) are identical.
	{The same action can be performed to satisfy multiple subtasks on the left, which improves the overall
		efficiency. Furthermore, all subtasks on the right satisfy the partial relations between
		the corresponding subtasks on the left, while additional constraints are added if there are conflicts
		among the ordering constraints. For instance, action $D^{\emptyset}_{w_7,w_7}$ of
		$\omega_{15}$ should not be executed before the subtask $\omega_{14}$}.
	Thus, an additional ordering constraint is added
	such that subtask $\omega_{15}$ should be executed after $\omega_{14}$. 
	These properties guarantee that each subtask within the R-posets on the left can be executed,
	as their partial relations are satisfied.
	Consequently, the final R-poset satisfies all R-posets on the left.
	Note that the complete formula has a total length of $62$,
	of which the NBA takes more than~$1h$ to compute.
	On the contrary, the first final R-poset is computed within $10.96s$.

	The results of task assignment are shown in the Fig.~\ref{fig:trajectory_gantt},
		which include the updates at $40,180,215,400s$ with~$5$ additional objects added.
		Then, after modifying the current R-posets to accommodate these formulas,
		the method TBCN is executed to assign the new subtasks within the final R-poset.
	The trajectories of agents and objects are shown,
	with certain regions that are not allowed to enter.
	For instance, when a patient vomits at region $h13$ in Fig.~\ref{fig:trajectory_gantt}~(c),
	the nurses cannot enter until other agents have cleaned this region.
	These constraints ensure that both their actions and their trajectories
	satisfy the R-poset.
	In addition, once a new object is added, a new formula is released and then the R-poset is updated.
	All subtasks satisfy the ordering constraints in the R-posets.
	{For instances, as shown in Fig.~\ref{fig:trajectory_gantt}(d),
		although agents $6,14$ have arrived in region~$w7$ before $100s$
		to collaborate on the subtask~$\sigma_{16}=\{M^{\emptyset}_{w7,w7}\}$  (in blue),
		they have to wait until that agent $10$ has fulfilled the subtask $\sigma_{14}=\{C^\emptyset_{w7,w7}\}$,
		due to the ordering constraints that~$(\omega_{14},\omega_{16})\in\preceq, \{\omega_{14},\omega_{16}\}\in\neq$.
		The constraints introduced by objects are also satisfied, e.g.,
		task $\omega_7$ (in green) cannot be executed at $380s$
		before subtask $\omega_6$ (in purple) has been finished,
		since the required object $3$ has not been transferred to region $o_4$.
		Last but not least, most tasks are executed in parallel mostly with
		a total completion time of $815ss$, which is much shorter than~$2013.5s$
		if the subtasks are executed sequentially.}
	
	\subsection{Scalability Analysis and Comparisons} \label{scale_analysis}

	\begin{figure*}[t]
		\begin{minipage}[t]{0.50\linewidth}
			\centering
			\includegraphics[width=1\linewidth]{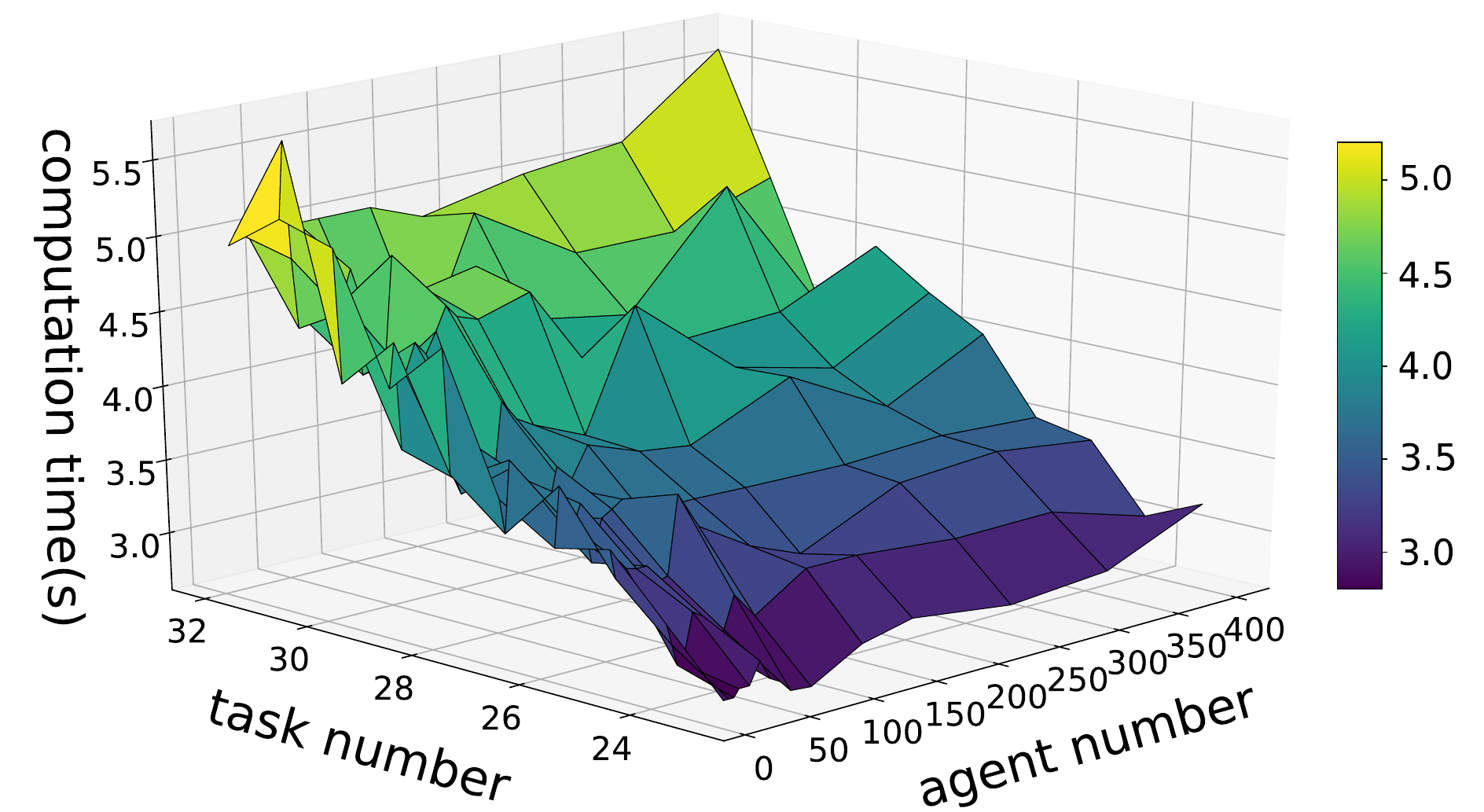}
			
		\end{minipage}%
		\begin{minipage}[t]{0.50\linewidth}
			\centering%
			\includegraphics[width =1\textwidth]{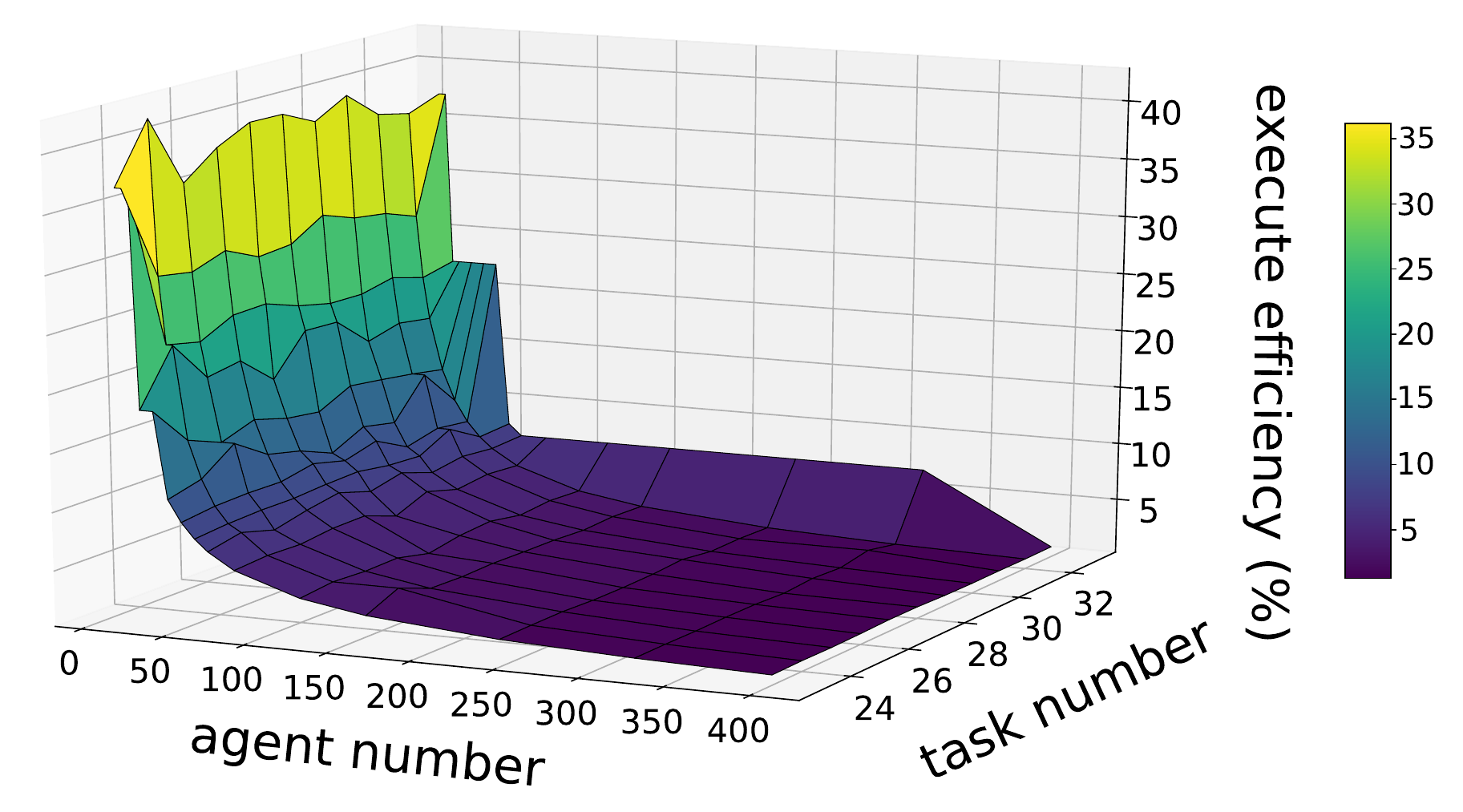}
		\end{minipage}%
		
		\caption{The computation time (\textbf{Left}) and the execution efficiency $\eta$ (\textbf{Right})
				with respect to different number of agents and subtasks. } 
		\label{fig:baseline_compare_scale}
		
	\end{figure*}
	\begin{figure*}[t] 
		\begin{minipage}[t]{0.48\linewidth}
			\centering
			\includegraphics[width=1\linewidth]{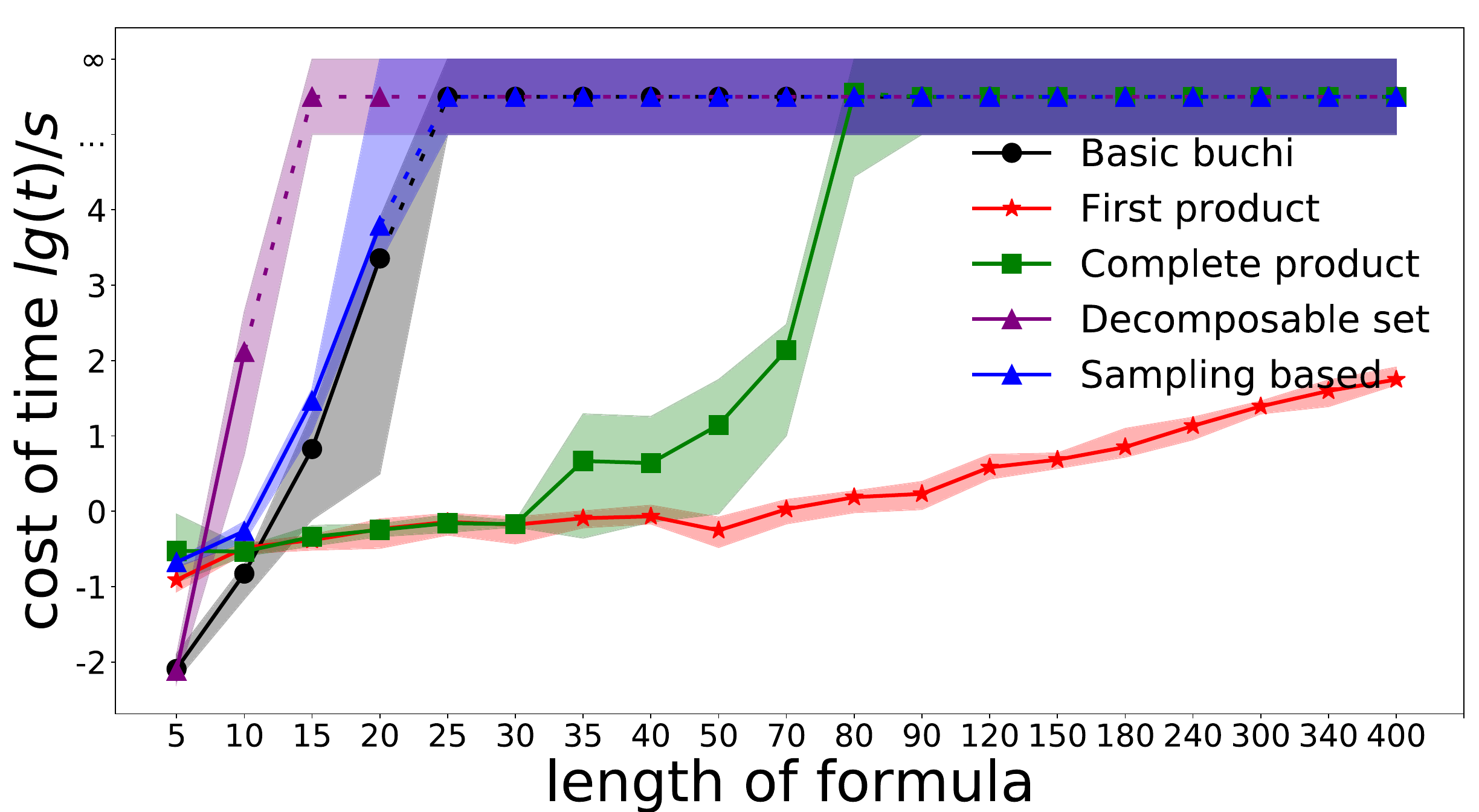}
			
		\end{minipage}%
		\begin{minipage}[t]{0.48\linewidth}
			\centering
			\includegraphics[width=1\linewidth]{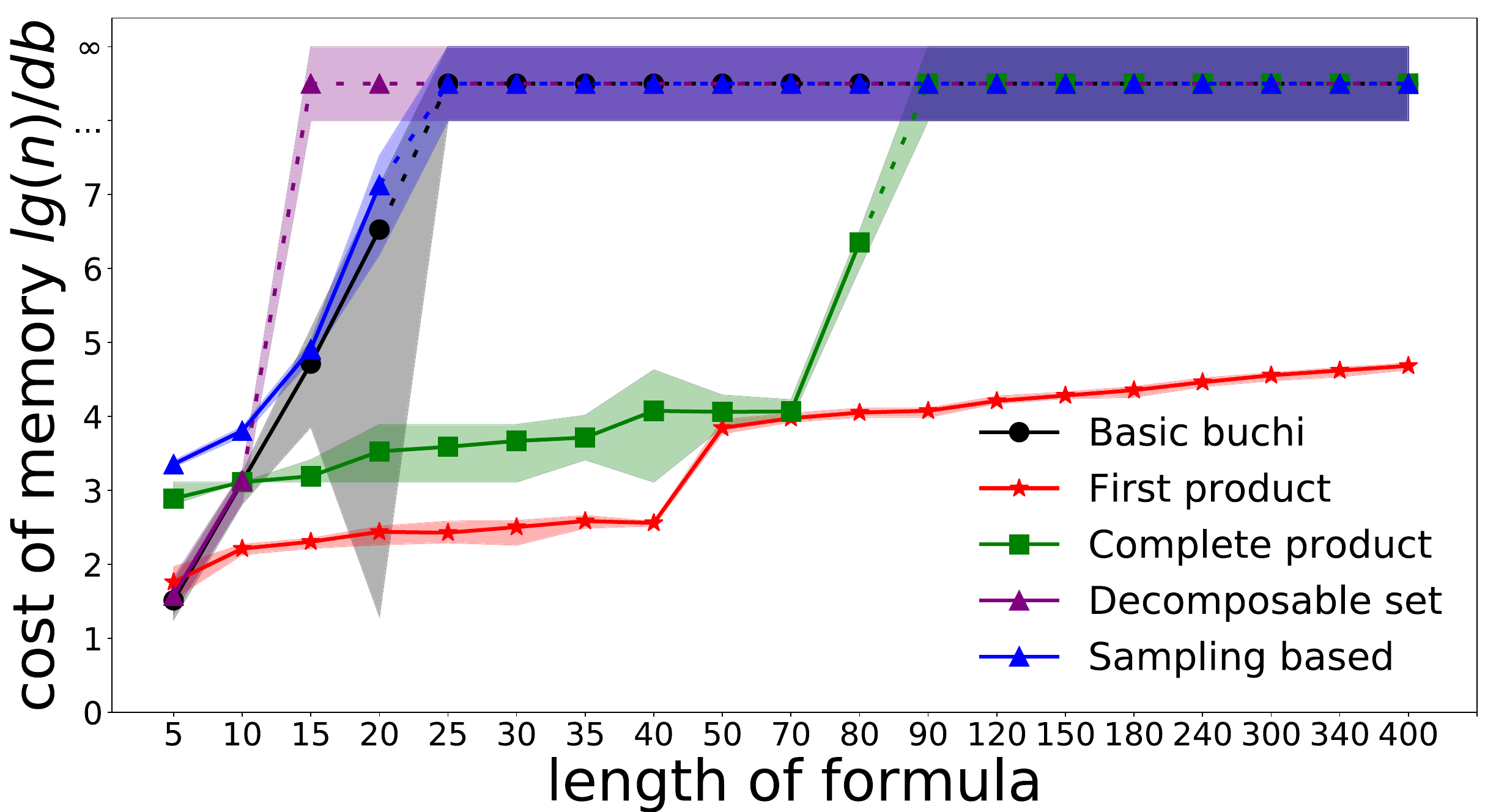}
			
		\end{minipage}%
		\caption{The comparison of the computation time (\textbf{Left}) 
				and memory (\textbf{Right}) by different methods. } 
		\label{fig:baseline_compare_others}
		
	\end{figure*}

 	{First, we show how the computation time of the proposed task assignment method TBCN
		varies under different numbers of agents and subtasks.
		Since the number of subtasks cannot be directly determined, we run the TBCN
		with a large number of formulas, of which the length ranges from $20$ to $80$.
		The number of subtasks and the associated computation time are recorded.
		As shown in Fig.~\ref{fig:baseline_compare_scale},
		the average computation time only increases slightly as the number of agents
		is increased from $12$ to $400$, while the computation time increases considerably if the number
		of subtasks is increased from $22$ to $34$.
		This is due to the fact that new subtasks would introduce additional temporal
		constraints in the assignment procedure.
		The execution efficiency $\eta$ from \eqref{eta_of_efficiency}
		decreases as the number of agent increases,
		while $\eta$ increases slightly as the number of subtasks grows.
		The highest $\eta$ is about $39\%$ where most subtasks are executed in parallel
		with only minimum waiting time for task synchronization.}

	{Secondly, to further validate the scalability of the proposed method against existing methods,
		we evaluate the time and memory cost to compute
		both the first R-poset and the complete R-posets by the proposed \emph{Poset-Prod},
		given the task formulas of different lengths.
		The conversion from a LTL formula to NBA is via \texttt{LTL2BA}.
		The following three baselines are considered:
		(i) the direct translation from the complete formula to NBA~\cite{10.1007/3-540-44585-4_6}, in
		which the complete formula is the conjunction of all subformulas;
		(ii) the decomposition-set-based algorithm~\cite{faruq2018simultaneous}, which
		decomposes the NBA into independent subtasks;
		(iii) the sampling-based method~\cite{kantaros2020stylus,kantaros2018sampling}, which generates a product
                  automaton much smaller than the complete one by sampling the
                  product states of NBA and WTSs via RRT.
		Each method is tested three times with five formulas of the same length
		ranging from $5$ to $400$.
		As shown in Fig.~\ref{fig:baseline_compare_others},
		both the time and memory cost increase drastically
		with the formula length for all methods except the proposed \emph{Poset-prod}
		to compute the first R-poset.
		In particular, when the formula length exceeds $15$, the decomposition algorithm runs out of memory or time.
	The sampling-based method can not generate a solution when the formula length exceeds $20$.
			Since all the baseline methods require the translation to NBA first,
			it becomes intractable as the formula length exceeds $25$.
			In contrast, the proposed method of \emph{Poset-prod} can generate all R-posets with a formula length of $70$
			and the first R-poset even when the length reaches $400$,
			which is consistent with our analyses in Sec.~\ref{discussion}.}

	\subsection{Hardware Experiment}
	\label{hardware_experiment}
	
	\begin{figure*}[t]
		\begin{minipage}[t]{0.25\linewidth}
			\centering
			\includegraphics[width=1\linewidth]{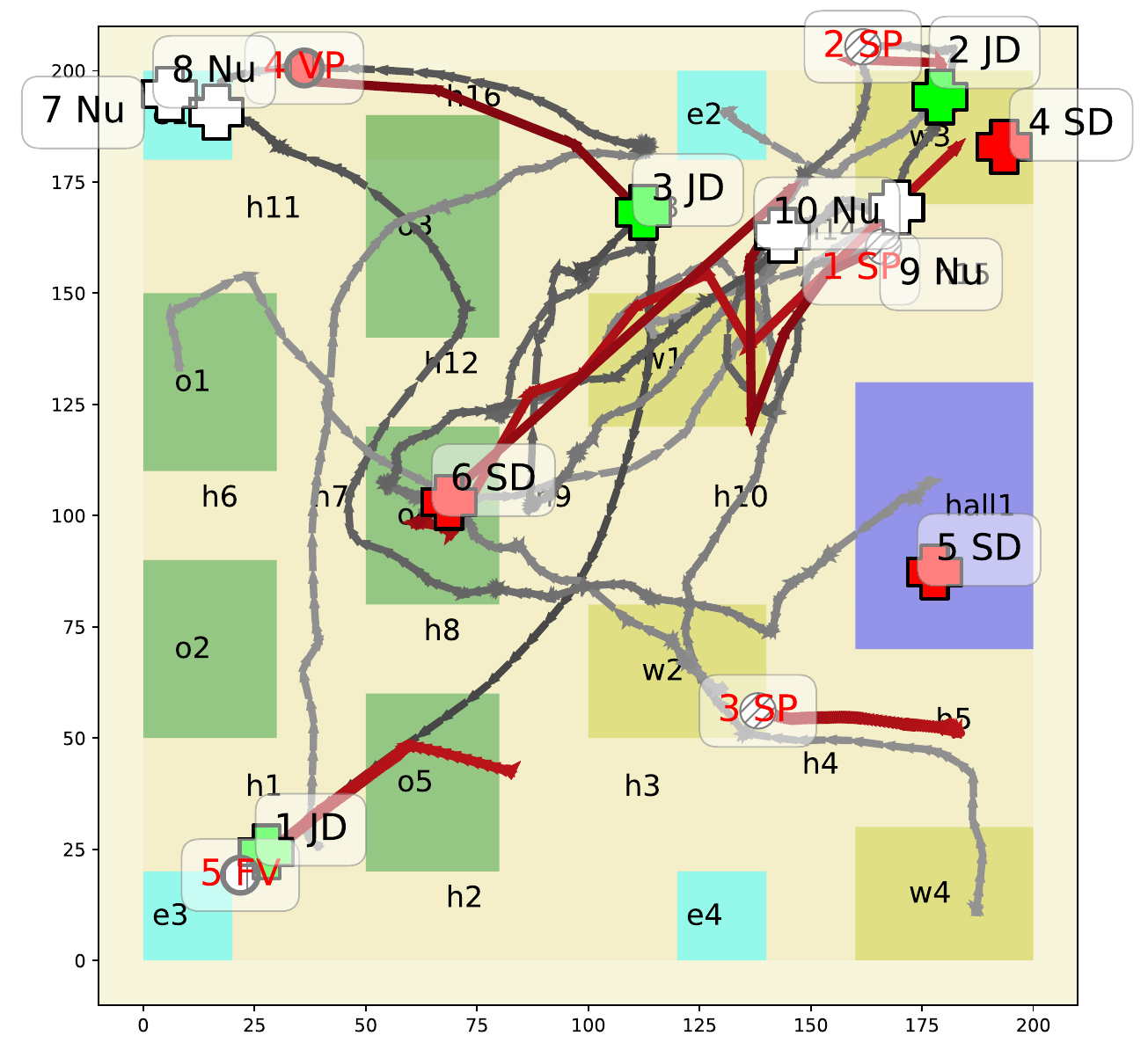}
			\label{fig:experiments(a)}
		\end{minipage}
		\begin{minipage}[t]{0.23\linewidth}
			\centering
			\includegraphics[width=1\linewidth]{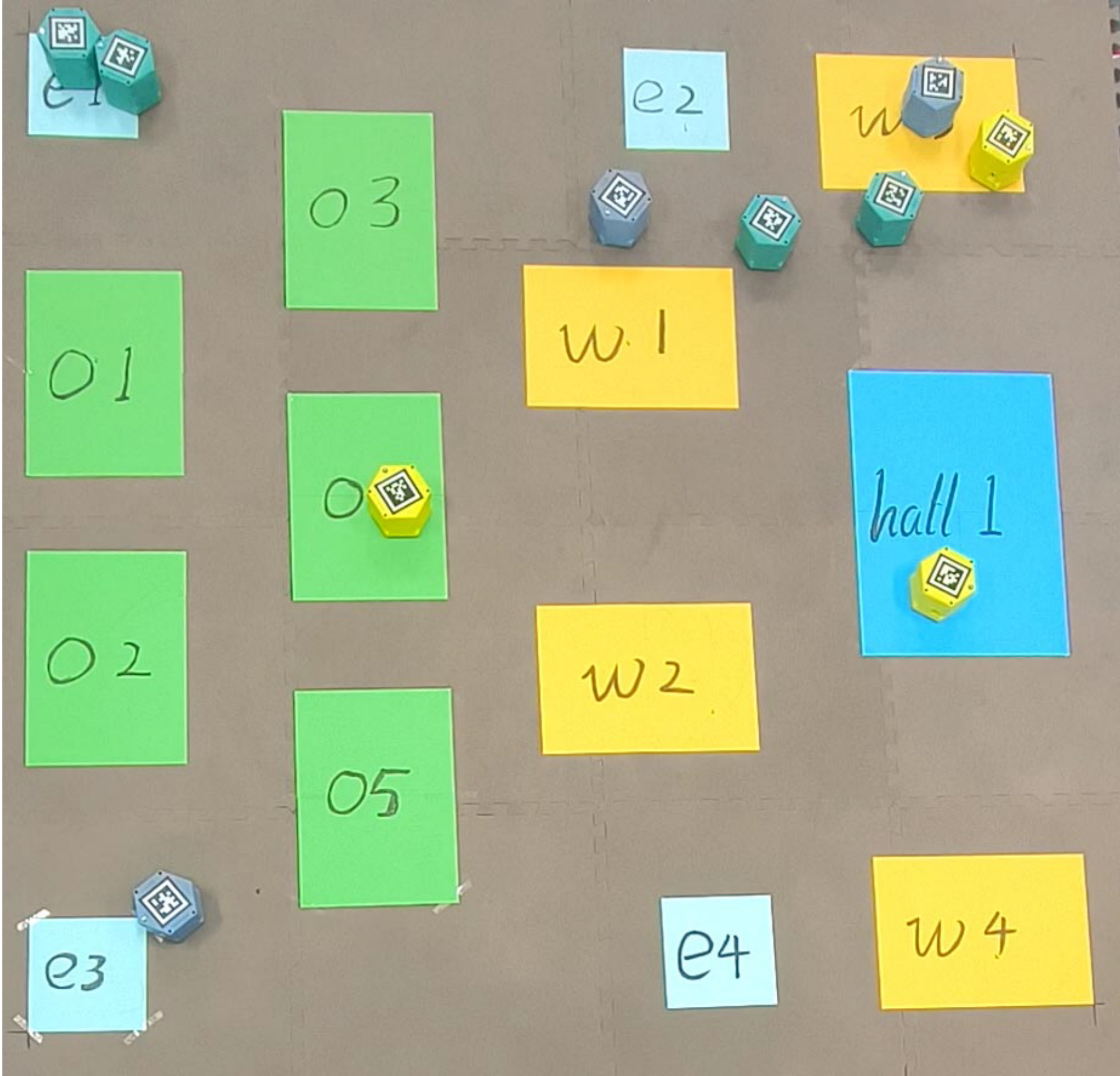}
			\label{fig:experiments(b)}
		\end{minipage}
		\begin{minipage}[t]{0.51\linewidth}
			\centering
			\includegraphics[width=1\linewidth]{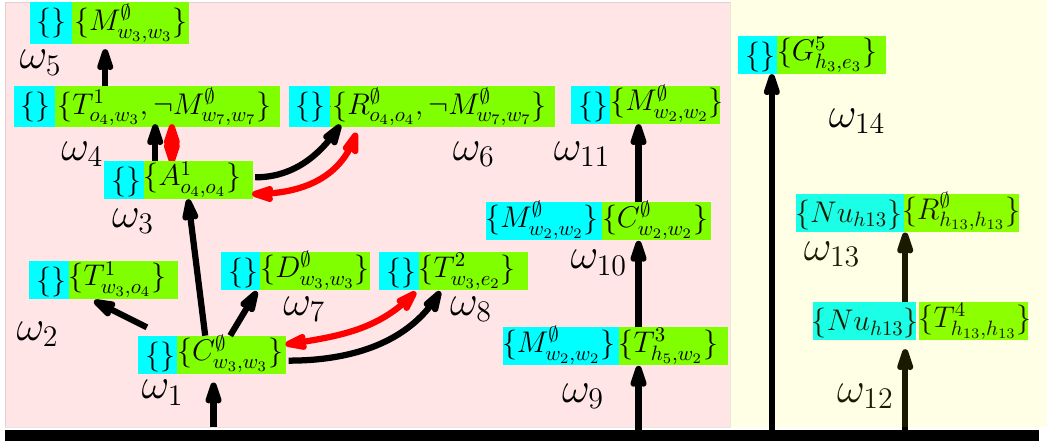}
			\label{fig:experiments(c)}
		\end{minipage}%
		\caption{The trajectories of agents in experiment (\textbf{Left}), experimental snapshot (\textbf{Middle})
				and the final R-poset (\textbf{Right}).}
		\label{fig:experiments}
		
	\end{figure*}

	\begin{figure*}[t] 
		\begin{minipage}[t]{0.50\linewidth}
			\centering
			\includegraphics[width=\linewidth,height=0.4\linewidth]{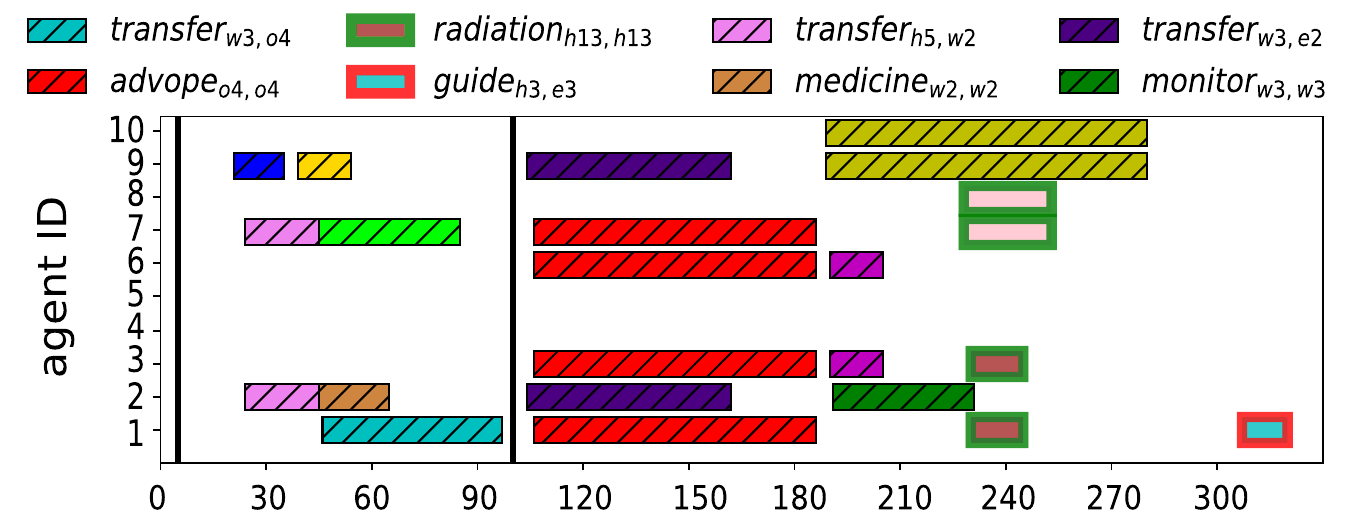}
			\label{fig:experiments(d)}
		\end{minipage}
		\begin{minipage}[t]{0.49\linewidth}
			\centering
			\includegraphics[width=\linewidth,height=0.42\linewidth]{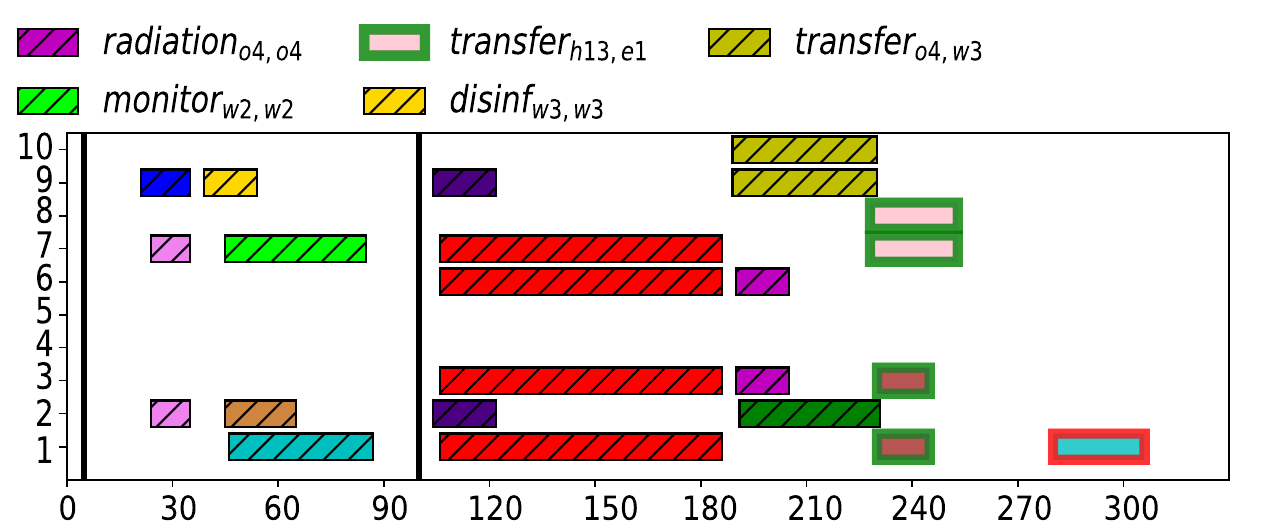}
			\label{fig:experiments(e)}
		\end{minipage}	
		\caption{The Gantt graph of the planned execution (\textbf{Left}) 
				and the Gantt graph of the actual execution (\textbf{Right}).}
		\label{fig:experiments2}		
	\end{figure*}
	
	We further tested the proposed method on hardware within a simplified hospital environment.
	The multi-agent system consists of $10$ differential-driven mobile robots,
	with 3 $JD$ in green, 3 $SD$ in yellow and 4 $Nu$ in blue.
	The required tasks include
	"prepare and execute an advanced operation for patient $1$ at operating room $o4$".
	Similar to Sec.~\ref{simulation}, there are event-triggered tasks
	released online such as ``when a patient vomits, take him to the ward,
	do not enter this region until it is cleaned." The exact task description
	and formulas are shown in Table.~\ref{table:definition_formulas_experiment}.
	
	{In summary, the system is required with $6$ sub-formulas, and the final formula is  $\varphi=\varphi_{b1}\land\cdots\land\varphi_{b_4}\land\varphi_{FV}\land\varphi_{JP}$, with the total length $27$,
		thus can not be translated into a NBA directly.
		The agent trajectories within $185s$ are shown in Fig.~\ref{fig:experiments}.
		As shown in the right of Fig.~\ref{fig:experiments}, the final R-poset consists of~$14$ subtasks,
		where the part in red is calculated offline and the part in yellow is generated online.
		Most subtasks are executed in parallel, meaning that the R-posets can be satisfied with a high efficiency.
		The complete R-poset product is derived in $3.1s$ and the task assignment method TBCN is activated three times,
		of which the average planing time is~$2.7s$.
		As shown in the left of Fig.~\ref{fig:experiments2}, the Gantt graph is updated twice at $5s$ and $100s$
		during execution, and subtasks that are released online are marked by green boxes.
		It is worth noting that the agent movement during real execution
		requires more time due to motion uncertainty, communication delay, drifting and collision avoidance.
		Nonetheless, the proposed method can adapt to these fluctuations and still satisfy the specified tasks,
		as shown in the Gantt graph of the overall execution in Fig.~\ref{fig:experiments2}.
		Experiment videos are provided in the supplementary files.}

\section{DISCUSSION}
\label{discussion}
In this work, we propose \emph{Poset-prod}, an efficient online task planning algorithm
for multi-agent systems
where tasks are released dynamically and constantly online.
It consists of a systematic method to convert the temporal tasks into their equivalent R-posets,
of which their products are computed online.
Given these R-posets, an anytime task assignment algorithm is proposed to adapt
the local plans of robots online, such that the overall safety and correctness is ensured.
The overall framework is shown to be fast and efficient, thus particularly suitable for large-scale
multi-agent systems collaborating in dynamic environments. 

The most significant advantage of \emph{Poset-prod} is that the
complexity of obtaining the first solution only grows linearly with the length of formulas.
Given a set of formulas~$\{\varphi_{i},i\leq m\}$ and
$\max_i|\varphi_{i}|=n$, deriving the first solution has a polynomial complexity of~$O(m^2n^3)$.
Despite that the overall complexity to compute the complete R-posets of $\{\varphi_i\}$ is $O(n^{5m} )$,
it is still much smaller than the complexity of calculating the NBA
of the conjunction~$\varphi$ via \texttt{LTL2BA}~\cite{10.1007/3-540-44585-4_6}, which is $O(mn\cdot 2^{mn})$.
Thus, our method can plan for task formulas with a length of about~$400$ within about $50s$,
while most existing methods fail at the formula length of~$25$.

Future work involves two directions:
(i) extending the sc-LTL task formulas
	to general LTL and other languages such as CTL~\cite{koymans1990specifying} and STL \cite{maler2004monitoring}.
	It remains unclear how general LTL formulas with \emph{always} operators
	can be incorporated in the R-poset, especially with the prefix-suffix
	structure;
(ii) considering unknown and uncertain environments
	that are modeled as Markov Decision Processes (MDP).
	In this case, a reactive high-level plan is essential
	to take into account all possible environment behaviors.

\section{MATERIALS AND METHODS}
\label{sec:method}
In this section, we provide the knowledge of LTL in Preliminaries, 
the definition of alphabets and objective function in problem formulation,
and algorithm details in Approach.

\subsection{Preliminaries} 
\label{subsec:LTL}

As mentioned in Sec.\ref{Result_taskspecification} of Result, the basic ingredients of Linear Temporal Logic (LTL) formulas are a set of atomic propositions $AP$
in addition to several Boolean and temporal operators.  
For a given LTL formula $\varphi$,
there exists a Nondeterministic B\"{u}chi Automaton (NBA) as follows:
\begin{definition}\label{def:nba}
	A NBA $\mathcal{A}\triangleq (S,\,\Sigma,\,\delta,\,(S_0,\,S_F))$
	is a 4-tuple, where~$S$ are the states;
	$\Sigma=AP$;
	$\delta:S\times \Sigma\rightarrow2^{S}$ are transition relations;
	$S_0, S_F\subseteq S$ are initial and {accepting} states.  
\end{definition}
An infinite {word} $w$ over the alphabet $2^{AP}$ is defined as an
infinite sequence $W=\sigma_1\sigma_2\cdots, \sigma_i\in 2^{AP}$.
The language of $\varphi$ is defined as the set of words that satisfy $\varphi$,
namely, $\mathcal{L}(\varphi)=Words(\varphi)=\{W\,|\,W\models\varphi\}$ and $\models$ is the satisfaction relation.
Additionally, the resulting \emph{run} of~$w$ within~$\mathcal{A}$
is an infinite sequence~$\rho=s_0s_1s_2\cdots$
such that $s_0\in S_0$, and $s_i\in S$, $s_{i+1}\in\delta(s_i,\,\sigma_i)$ hold for all index~$i\geq 0$.
A run is called \emph{accepting} if it holds that
$\inf(\rho)\cap {S}_F \neq \emptyset$,
where $\inf(\rho)$ is the set of states that appear in $\rho$ infinitely often.
{A special class of LTL formula called \emph{syntactically co-safe} formulas (sc-LTL)~\cite{Katoen2008Principles, belta2017formal},
	which can be satisfied by a set of finite sequence of words.
	They only contain the temporal operators $\bigcirc$, $\textsf{U}$ and $\Diamond$
	and are written in positive normal form where the negation
	operator $\neg$ is not allowed before the temporal
	operators.}
A relaxed partially ordered set (R-poset) over an NBA $\mathcal{B}_\varphi$ is defined as follows:

\begin{definition}[R-poset]~\cite{LIU2024111377}
	R-poset is a 3-tuple: 
	$P_\varphi=(\Omega_\varphi,\preceq_\varphi,\neq_\varphi)$:
	$\Omega_\varphi=\{(\ell,\sigma_{\ell} , \sigma^{s}_\ell),\forall\ell=0,\dots,L\}$ is the set of {subtasks},
	where $\ell$ is the index of subtask $\omega_\ell$;
	$\sigma_{\ell}\subseteq\Sigma$ are the transition labels;
	$\sigma^{s}_\ell\subseteq\Sigma$ are the self-loop labels from Def.~\ref{def:nba}~.
	$\preceq_{\varphi}\subseteq \Omega_{\varphi} \times \Omega_{\varphi}$ is the "less equal"
	relation:
	If $(\omega_h,\omega_{\ell}) \in \preceq_{\varphi}$ or
	equivalently $\omega_h\preceq_{\varphi}\omega_{\ell}$,
	then $\omega_{\ell}$ can only be \emph{started} after $\omega_h$ is started.
	$\neq_{\varphi}\subseteq 2^{\Omega_{\varphi}}$ is the "opposed" relation: If
	$\{\omega_h,\cdots,\omega_\ell\}\in \neq_{\varphi}$ or
	equivalently $\omega_h \neq_\varphi\cdots\neq_\varphi\omega_\ell$, subtasks in~$\omega_h,\cdots,\omega_\ell$
	cannot all be executed simultaneously.
\end{definition}

A word is \emph{accepting} if
it satisfies all the constraints imposed by~$P_{\varphi}$.
Additionally, the word of R-poset will 
satisfy the NBA as $Words(P_{\varphi}) \subset Words(\varphi)$.

\begin{definition}[Language of R-poset]\cite{LIU2024111377}
	\label{def:language}
	Given a word $w=\sigma'_1\sigma'_2\cdots$ satisfying R-poset $P_\varphi=(\Omega_\varphi,\preceq_\varphi,\neq_\varphi)$, denoted as $w\models P_\varphi$, it holds
	that: i) given $\omega_{i_1}=(i_1,\sigma_{i_1},\sigma^s_{i_1})\in\Omega_\varphi$, there exist $j_1$ with
	$\sigma_{i_1}\subseteq\sigma'_{j_1}$ and $ \sigma^s_{j_1}\cap\sigma'_{m_1}=\emptyset, \forall m_1<j_1$;
	ii) $\forall (\omega_{i_1},\omega_{i_2})\in\preceq_\varphi, \exists j_1\leq j_2,
	\sigma_{i_2}\subseteq\sigma'_{j_2}, \forall m_2< {j_2},\sigma^s_{j_2}\subseteq\sigma'_{m_2}$;
	iii) $\forall (\omega_{i_1},\cdots,\omega_{i_n})\in\neq_\varphi,
	\exists \ell\leq n, \sigma_{i_\ell}\not\subseteq\sigma'_{j_1}$.
	Language of R-poset $P_\varphi$ is the set of all word $w$ that satisfies $P_\varphi$,
	denoted by $\mathcal{L}(P_\varphi)\triangleq \{ w|w\vDash P_\varphi\}$. 
\end{definition}

Assuming that $\mathcal{P}_{b_i}=\{P^{b_i}_1,P^{b_i}_2,\cdots\}$
is the set of all possible R-posets of NBA $\mathcal{B}_{b_i}$, it is shown in Lemma 3 of our previous work~\cite{LIU2024111377} that:
(i) $\mathcal{L}(P^{b_i}_j)\subseteq \mathcal{L}(\mathcal{B}_{b_i}), P^{b_i}_j\in \mathcal{P}^{b_i}_\varphi$;
(ii) $\bigcup_{P^{b_i}_j\in \mathcal{P}^{b_i}_\varphi}\mathcal{L}(P_i)=\mathcal{L}(\mathcal{B}_{b_i})$.
In other words, the R-posets contain the necessary information for subsequent steps.

\begin{figure*}[!t]
	\centering
	\includegraphics[width=1\linewidth]{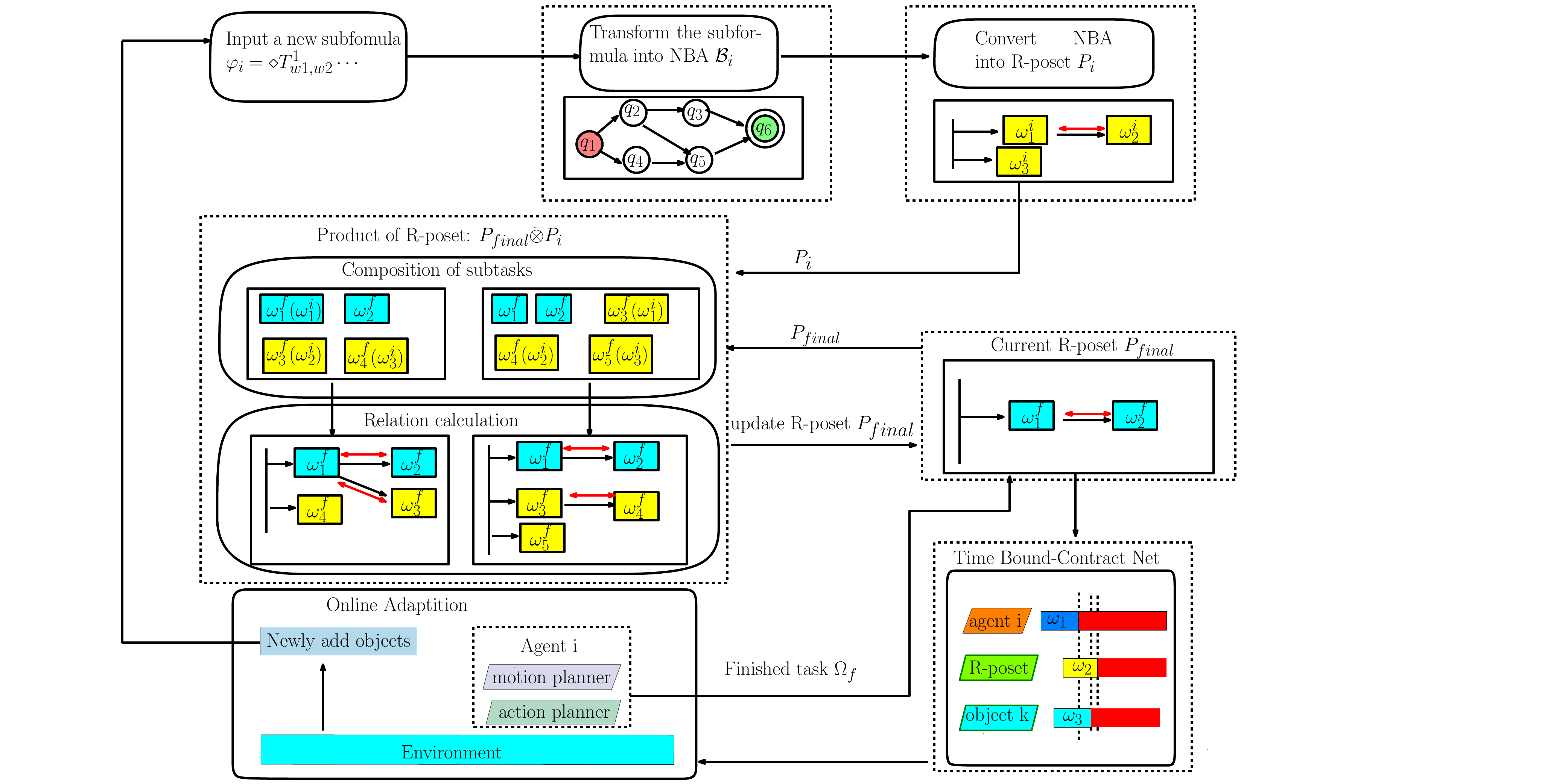}
	\vspace{-0.3in}
	\caption{Framework of the proposed method. For a new input formula $\varphi_i$, we firstly change it into an R-poset $P_i$ with the method in Sec.~\ref{subsec:LTL}.
		Then, \emph{Poset-prod} between $P_i$ and $P_{final}$ is calculated as in Sec.~\ref{Product_Poset}.
		Thirdly, after the R-poset $P_{final}$ is updated, the TBCN method in Sec.~\ref{Task_Assignment} determines and adjusts the task sequence of each agent and object. 
	New tasks are released when the agents execute 
			their local plans and detect new objects online. }
	\label{fig:logic}
	\vspace{-0.1in}
\end{figure*}

\subsection{Problem Formulation}\label{sec:problem}

\subsubsection{Collaborative Multi-agent Systems}\label{subsec:multi-agent}
Consider a workspace $W\subset \mathbb{R}^2$ with $M$ regions of interest
denoted as~$\mathcal{W}\triangleq \{W_1,\,\cdots,W_M\}$,
where~$W_m\in W$.
Furthermore, there is a group of agents denoted by~$\mathcal{N}\triangleq \{1,\cdots,N\}$
with different types $\boldsymbol{L}\triangleq \{1,\cdots,L\}$.
More specifically, each agent $n\in \mathcal{N}$ belongs to only one type $l=M_{type}(n)$,
where $M_{type}: \mathcal{N}\rightarrow\boldsymbol{L}$.
Each type of agents~$l \in \boldsymbol{L} $ is capable of providing a set of different actions denoted by~$\mathcal{A}_l$.
The set of all actions is denoted as $\mathcal{A}_a=\bigcup_{l\in L}\mathcal{A}_l=\{a_1,\cdots,a_{n_\mathcal{C}}\}$.
Without loss of generality, the agents can navigate between each region via the same transition graph,
i.e., $\mathcal{G}=(\mathcal{W},\,\rightarrow_\mathcal{G},\,d_\mathcal{G})$,
where~$\rightarrow\subseteq \mathcal{W}\times \mathcal{W}$
represents the allowed transitions; and
$d_\mathcal{G}:\rightarrow_\mathcal{G}\rightarrow \mathbb{R}_+$ maps each transition to
its duration.

Moreover, there is a set of interactive objects~$\mathcal{O}\triangleq\{o_1,\cdots,o_U\}$
with several types $\mathcal{T}\triangleq\{T_1,\cdots,T_H\}$
scattered across the workspace $W$.
These objects are \emph{interactive} and can be transported by the agents from one region to another.
An interactive object~$o_u\in\mathcal{O}$ is described by a three-tuple:
\begin{equation}\label{eq:tup}
o_u\triangleq(T_{h_u},\, t_{u},\, W^p_{u}),
\end{equation}
where $T_{h_u}\in \mathcal{T}$ is the type of object;
$t_u\in R^+$ is the time when $o_u$ appears in workspace $W$;
$W^p_{u}:R^+\rightarrow \mathcal{W}$ is a function that $W^p_{u}(t)$ returns the region of $o_u$ at time $t \geq t_u$;
and $W^p_{u}(t_u)\subseteq W$ is the initial region.
Additionally, new objects appear in $W$ over time and are then added to the set~$\mathcal{O}$.
With a slightly abusive of notation, we denote
the set of initial objects $\mathcal{O}_{in}$ that already exist in $W$
and the set of online objects $\mathcal{O}_{on}$ that are added during execution,
i.e., $\mathcal{O}=\mathcal{O}_{in}\cup\mathcal{O}_{on}$.

To interact with the objects, the agents can provide a series of collaborative
behaviors~$\mathcal{C}\triangleq\{C_1,\dots,C_K\}$.
A \emph{collaborative} behavior $C_k\in\mathcal{C}$ is a tuple defined as follows:
\begin{equation}\label{eq:c-k}
C_k\triangleq (o_{u_k},\{(a_{i},n_{i}),\, i\in \boldsymbol{L}_k\})
\end{equation}
where~$o_{u_k}\in\mathcal{O} \cup \{\emptyset\}$ is the interactive object if any;
$a_i\in\mathcal{A}_a$ is the set of cooperative actions required;
$0<n_{i}\leq N$ is the number of agents to provide the action $a_i$;
and~$\boldsymbol{L}_k$ is the set of action indices associated with the behavior~$C_k$.
Also, $d_k$ denotes the execution time of $C_k$.

A behavior can only be executed if the required object
is at the desired region.
Since objects can only be transported by the agents,
it is essential for the planning process to find the correct order of
these transportation behaviors.
Related works~\cite{verginis2022planning, 2018Multi} build a transition system to
model the interaction between objects and agents,
the size of which grows exponentially with the number of agents and objects.


\subsubsection{Task Specifications}\label{subsec:task-specification}
Consider the following two types of atomic propositions:
(i) $p^l_{m}$ is \emph{true} when any agent of type $l\in \boldsymbol{L}$ is in region
${W}_m\in {\mathcal{W}}$; $p^r_{m}$ is \emph{true} when any object of type $r\in \mathcal{T}$
is in region ${W}_m\in {\mathcal{W}}$;
Let $\mathbf{p}\triangleq \{p^l_{m},\, \forall {W}_m \in {\mathcal{W}}, l \in L\}
\bigcup \{p^r_{m},\, \forall {W}_m \in {\mathcal{W}}, r \in \mathcal{R}\}$.
(ii) $(C_k)^{{u_k}}_{k_1,k_2}$ is \emph{true} when the collaborative behavior
$C_k$ in~\eqref{eq:c-k} is executed with object~$o_{u_k}$, starting from region~${W}_{k_1}$ and
ending in region~${W}_{k_2}$.
Let $\mathbf{c} \triangleq\{(C_k)^{u_k}_{k_1,k_2},\forall C_k \in \mathcal{C},o_{u_k}\in\mathcal{O},\forall{W}_{k_1},
{W}_{k_2}\in {\mathcal{W}}\}$.
Given these propositions, the team-wise task specification is
specified as a sc-LTL formula
over~$\{\mathbf{p},\mathbf{c}\}$:
\begin{equation}\label{eq:updatetask}
\varphi=(\bigwedge_{i\in \mathcal{I}}\varphi_{\texttt{b}_i}) \wedge (\bigwedge_{o_u\in \mathcal{O}_{on}} \varphi_{\texttt{e}_u}),
\end{equation}
where $\{\varphi_{\texttt{b}_i},\,i=1,\cdots,I\}, \{\varphi_{\texttt{e}_u},\,o_u\in \mathcal{O}_{on}\}$
are two sets of sc-LTL formulas over~$\{\mathbf{p}, \mathbf{c}\}$. The $\varphi_{\texttt{b}_i}$
is specified in advance while $\varphi_{\texttt{e}_u}$ is generated online
when a new object $o_u$ is added to $\mathcal{O}_{on}$. 

To satisfy the LTL formula $\varphi$, the complete action sequence of all agents is defined as
\begin{equation}
\begin{aligned}
\mathcal{J}=(J_1,J_2,\cdots,J_N),
\end{aligned}
\label{Initial_solution}
\end{equation}
where $J_n\in\mathcal{J}$ is the sequence of~$(t_k, \mathcal{C}_k, a_k)$,
which means that agent $n$ executes behavior $\mathcal{C}_{k}$
by providing the collaborative service $a_k\in \mathcal{A}_a$ at time $t_k$.
In turn, the sequences of actions for an interactive object is defined as:
\begin{equation}
\begin{aligned}
\mathcal{J}^o=(J^o_1,J^o_2,\cdots,J^o_U),
\end{aligned}
\label{Initial_solution_resource}
\end{equation}
where $J^o_u=(t_k,\mathcal{C}_{k})\cdots$ is the sequence of tasks
associated with object $o_u\in\mathcal{O}$.
The task pair $(t_k,\mathcal{C}_{k})$ is added to $J^o_u$ if
object $o_u$ joins behavior $\mathcal{C}_k$ at time $t_k$. 
Assume that the duration of formula~$\varphi_{\texttt{b}_i}, \varphi_{\texttt{e}_i}$ from being published to being satisfied is given by~$D_i$,
the average efficiency is defined as
\begin{equation}
\begin{aligned}
\eta\triangleq \frac{\sum_{C_k\in J} |\boldsymbol{L}_k|d_k}{D_i},
\end{aligned}
\label{eta_of_efficiency}
\end{equation}
which is the percentage of time when actions are performed.


\begin{problem}\label{prob:formulation1}
	Given the sc-LTL formula $\varphi$ defined in \eqref{eq:updatetask},
	synthesize and update the motion and action
	sequence of agents $\mathcal{J}$ and objects $\mathcal{J}_o$ for each agent $n\in \mathcal{N}$ to satisfy $\varphi$
	and maximize execution efficient $\eta$. 
\end{problem}

Although maximizing the task efficiency of a multi-agent system is a classical problem,
the combination of
interactive objects, long formulas and contingent tasks
imposes new challenges in terms of exponential complexity~\cite{10.1007/3-540-44585-4_6,2018Multi} and online adaptation~\cite{6942756}.

\subsection{Approach}
As shown in Fig.~\ref{fig:logic}, when a new R-poset is generated,
the proposed solution realizes the requirement through two main components:
i)  Product of R-posets, where the product of existing R-posets is computed incrementally;
ii) Task assignment, where subtasks are assigned to the agents
given the temporal and spatial constraints specified in the R-poset.

\subsubsection{Product of R-posets }\label{Product_Poset}
As the first two steps showed in Fig.~\ref{fig:logic}, 
when a new formula $\varphi_i\in \varPhi_{\texttt{b}} ,\varPhi_{\texttt{u}}$ is added, 
it will be transform into NBA first. Then, an R-poset $P_i$ is generated by the method
proposed in our previous work \cite{LIU2024111377}. 
The other R-poset $P_{final}$ involved in calculation is the previous result of \emph{Poest-prod}, which will be updated after this round calculation.
With $P_{final}$ as $P_1=(\Omega_{1},\preceq_{1},\neq_{1})$, $P_i$ as $P_2=(\Omega_{2},\preceq_{2},\neq_{2})$, we define product of R-posets as follows:
\begin{definition}[Product of R-posets]\label{def:Product_of_R_posets}
	Given a finite word $w_0$, the product of two R-posets $P_1,P_2$ is defined
	as a set of R-posets $\mathcal{P}^r=\{P'_1,P'_2,\cdots\}$, denoted as $\mathcal{P}^r=P_1\bigotimes P_2$ where $P'_i$ satisfies two conditions:
	(i) if $w_0w\in \mathcal{L}(P_1), w\in \mathcal{L}(P_2)$, then $w_0w \in\bigcup_{P'_i\in\mathcal{P}^r} \mathcal{L}(P')$;
	(ii) if $w_0w\in \mathcal{L}(P'_i)$, $P'_i\in \mathcal{P}^r$, then $w_0w\in \mathcal{L}(P_1), w\in \mathcal{L}(P_2)$.
	
\end{definition} 
The word~$w_0$ is already executed, containing the finished subtasks $\Omega^1_{finish}$ of $P_1$.
Thus, $w_0$ can not influence $P_2$ whose subtasks will be executed in the future.
Specially, if the new formula is offline as $\varphi_i\in\varPhi_{\texttt{b}}$ and 
the agents have not started to execute subtasks, $w_0$ will be set as empty. 
As showed in Fig.~\ref{fig:logic}, \emph{Poset-prod} consists of following two steps.

(i) \textbf{Task Composition}.
In this step, we generate all possible combinations of subtasks which can satisfy
both $\Omega_1$ and $\Omega_2$.
Namely, a set of subtasks~$\Omega'=\{\omega'_1,\cdots,\omega'_{n'}\}$ satisfies $\Omega_1$
if for each $\omega^1_i=(i,{\sigma^1_i},{\sigma^{s1}_i})\in\Omega_1$,
there is a subtask $\omega'_j=(j,{\sigma_j}',{\sigma^{s}_j}')\in\Omega'$ satisfying
$\sigma^1_i \subseteq \sigma'_j$ and
$\sigma^{s1}_i\subseteq{\sigma^{s}_j}'$,
or $\omega'_j\models\omega^1_i$ for brevity.
Thus, we set $\Omega'=\Omega_1$ firstly and
$\Omega'$ clearly satisfies $\Omega_1$ with $\forall\omega^1_i\in\Omega_1, \omega'_i\models\omega^1_i$.
Then, a mapping function $M_{\Omega}:\Omega_2\rightarrow\Omega'$ is defined to store
the satisfying relationship from $\Omega_2$ to $\Omega'$, and $\mathcal{D}(M_\Omega)\in\Omega_2$ is the domain
of $M_\Omega$, $\mathcal{R}(M_\Omega)$ is the range of $M_\Omega$.
As all relations are unknown, $M_{\Omega}$, $\mathcal{D}(M_\Omega)$ and $\mathcal{R}(M_\Omega)$
are initially set to empty.
Namely, if $\omega'_j\models\omega^2_i$,
we will store $M_\Omega(\omega^2_i)=\omega'_j$ and $M^{-1}_\Omega(\omega'_j)=\omega^2_i$, and add $\omega^2_i$ into $\mathcal{D}(M_\Omega)$, $\omega'_j$ into $R(M_\Omega)$,
where $M^{-1}_\Omega$ is the inverse function of $M_\Omega$.

A depth-first-search (DFS) \cite{kozen1992depth} is used to add the subtasks of $\Omega_2$ to $\Omega'$ in order and these mapping relations will be recorded in $M_\Omega$.
The search sequences of DFS can be initialized as $que=[(\Omega'=\Omega_1,M_\Omega=\emptyset)]$.
Then, during the circle, we will fetch the first node of $que$ as $(\Omega',M'_\Omega)$. The next unmixed subtasks in $\Omega_2$
is $\omega^2_i,i=|M'_\Omega|+1$. Any subtask $\omega^1_j \in\Omega_1/ R(M'_{\Omega})/\Omega^1_{finish}$ with $\omega^1_{j}\models\omega^2_{i}$ or
$\omega^2_{i}\models\omega^1_{j}$ can create a new combination based 
on $(\Omega',M'_\Omega)$: 
\begin{equation}
\label{generate_new_subtask}
\begin{aligned}
\hat{\Omega}'&=\Omega',\, \hat{\omega}'_j=(j,\sigma^1_j\cup\sigma^2_i,\, \sigma^{s1}_j\cup\sigma^{s2}_i),\\
\hat{M}'_\Omega&=M'_\Omega,\, \hat{M}'_\Omega(\omega^2_i)=\omega'_j,
\end{aligned}
\end{equation}
which means the subtask $\omega'_j$ in $\Omega'$ can satisfy both $\omega^1_j,\omega^2_i$.
Moreover, for the subtask $\omega^2_i$, we can always create
a set of subtasks $\hat{\Omega}'$ and the corresponding mapping function $\hat{M}'_\Omega$ by appending
$\omega^2_i$ into $\hat{\Omega}'$ as~$\hat{\omega}'_j$
such that $\hat{\omega}'_j\models\omega^2_i$ holds,
i.e.,
\begin{equation}
\label{generate_new_subtask2}
\begin{aligned}
\hat{\Omega}'=\Omega',j=|\Omega'|+1, \hat{\omega}'_{j}=\omega^2_i;\\
\hat{M}'_\Omega=M'_\Omega, \hat{M}'_\Omega(\omega^2_i)=\omega'_{j},
\end{aligned}
\end{equation}
which means the subtask $\omega'_j$ can be executed to satisfy $\omega^2_i$. 
This step ends if the time budget $t_b$
or the search sequence $que$ exhausted.
Once $|D(M'_\Omega)|=|\Omega_2|$, a $\Omega'$ satisfying $\Omega_1,\Omega_2$ is already found.
In this case, the next step is triggered.
As showed in Fig.~\ref{fig:logic}~, one of found combination is 
$M^1_\Omega(1)=1, M^1_\Omega(3)=2, M^1_\Omega(4)=3$, and $\omega^f_1$
is created by~\eqref{generate_new_subtask}, and $\omega^f_3, \omega^f_4$
is created by~\eqref{generate_new_subtask2}.

(ii) \textbf{Relation Update}.
Given the set of subtasks
$\Omega'$ and the mapping function $M_\Omega$,
we calculate the partial relations among them and build a product R-poset $P$.
Firstly, we construct the ``less-equal'' constraint $\preceq$ as follows:
\begin{equation}\label{equ:inherit}
\begin{aligned}
\preceq=\preceq_1\cup \{(M_\Omega({\omega^2_{\ell_1}}),M_\Omega(\omega^2_{\ell_2}))|(\omega^2_{\ell_1},\omega^2_{\ell_2})\in \preceq_{2}\},
\end{aligned}
\end{equation}
which inherits both less equal relations $\preceq_1$ in $P_1$ and $\preceq_2$ in $P_2$.
Then, we update $\Omega'$ to consider the constraints imposed by
the self-loop labels in other subtasks.
Specifically, if a new relation $(\omega_i,\omega_j)$ is added to $\preceq$
by $(M^{-1}_\Omega(\omega^2_{i}),M^{-1}_\Omega(\omega^2_j))\in\preceq_2$ while
$(\omega^1_i,\omega^1_j)\not\in\preceq_1$ holds,
$\omega_i$ is required to
be executed before $\omega_j$ although $(\omega^1_i,\omega^1_j)$ does not belong to $\preceq_1$ of $P_1$.
In this case,  $\sigma_i$ and $\sigma^s_i$ are updated to guarantee
the satisfaction of the self-loop labels $\sigma^s_j$ before executing $\sigma_j$.
For each subtask $\omega_i$, the newly-added suf-subtasks from $\preceq_1,\preceq_2$ are defined as $S^1_p, S^2_p$,
i.e.,
\begin{equation}\label{new_partial_order}
\begin{aligned}
S^1_p=&\{\omega_j|(\omega_{i},\omega_{j})\in\preceq,(\omega^1_{i},\omega^1_j)\not\in \preceq_{1}\},\\
S^2_p=&\{M^{-1}_\Omega(\omega_j)|(\omega_{i},\omega_{j})\in\preceq,(M^{-1}_\Omega(\omega^2_i),M^{-1}_\Omega(\omega^2_j))\not\in \preceq_{2}\},\\
\end{aligned}
\end{equation}
where are the subtasks that should be executed after $\omega_i$.
Thus, the action labels~$\sigma_i$ and self-loop labels~$\sigma^s_i$ in
$\omega_i=(i,\sigma_i,\sigma^s_i)$ are updated accordingly as follows:
\begin{equation}
\label{omega_task_update}
\begin{aligned}
\hat{\sigma} =\bigcup_{\omega^1_\ell\in S^1_p}\sigma^{s1}_{\ell}\cup\bigcup_{\omega^2_\ell\in S^2_p}\sigma^{s2}_{\ell},
\sigma^s_i=\hat{\sigma} \cup\sigma^s_i,
\sigma_i=\hat{\sigma} \cup\sigma_i, 
\end{aligned}
\end{equation}
in which $\sigma^s_i$ and $\sigma_i$ should be executed under the additional labels $\hat{\sigma}$
thus the self-loop labels of $S^1_p, S^2_p$ are satisfied.

Finally, we find the potential ordering by checking whether a subtask $\sigma^s_i$ is
in conflicts with another subtask $\sigma_j$ while
$(\omega_i,\omega_j)\not\in\preceq$. 
If so, an additional ordering $(\omega_i,\omega_j)$ will be added to $\preceq$ as:
\begin{equation}
\label{neq_update}
\begin{aligned}
\preceq = \preceq \cup \{(\omega_i,\omega_j)|\sigma_j\not{\models}\sigma^s_i\}
\end{aligned}
\end{equation}
Then, $\Omega$ will be updated following~\eqref{new_partial_order} and \eqref{omega_task_update}.
Regarding the set of subtasks $\Omega$ that have no conflicts in $\preceq$,
its ``not-equal'' relations $\neq$ is generated by a simple combination as: 
\begin{equation}
\label{neq_calculate}
\begin{aligned} 
\neq=\neq_1\cup \Big\{\{{M_{\Omega}(\omega_{\ell_i})}\}| \{\omega_{\ell_i}\}\in \neq_{2}\Big\}. 
\end{aligned}
\end{equation}
The resulting poset $P_i=(\Omega,\preceq,\neq)$ is added to $\mathcal{P}_{final}$.

As shown in Fig.~\ref{fig:logic}, \textbf{Relation Update} gets two R-posets
and the partial relations of each R-poset are succeeded from $P_i, P_{final}$. 
Due to the anytime property,
the two steps procedure can be repeated until all possible R-posets are found
or just ended when the first R-poset is found.

\subsubsection{Task Assignment}
\label{Task_Assignment}
To satisfy the final R-poset $P_{final}=(\Omega_{f},\preceq_{f},\neq_{f})$, the subtasks
$\Omega_f$ should be executed under the partial orders of $\preceq_f, \neq_f$.
Each subtask $\omega_i\in\Omega_c$ represents a collaborative behavior $C_j$.
Thus, we can redefine the action sequence of each agent $J_n\in\mathcal{J}$
as $J_n=[(t_k,\omega_{k},a_k),\cdots]$ and the action sequence of each object $J^o_u\in\mathcal{J}_o$ as
$J^o_u=[(t_k,\omega_k, a_k),\cdots]$,
in which we replace the cooperative behavior $C_k$ with $\omega_k$ since $C_k\in\sigma_k$.

We propose a sub-optimal algorithm called Time Bound Contract Net (TBCN)
to generate and adapt the action sequence of agents and objects.
Compared with the classical Contract Net method \cite{1675516},
the main differences are: (i) the partial order $\preceq_f,\neq_f$ of R-poset might be changed
when new formula added, (ii) the assigned subtasks should satisfy the
partial orders in $\preceq_f,\neq_f$; (iii) the cooperative task should be fulfilled by multiple agents;
and (iv) interactive objects should be considered as an additional constraints.
TBCN solves these differences with three steps:
Firstly, we design a cancellation mechanism in the initialization to
adapt to the changes of R-poset mentioned in (i).
Secondly, the partial orders in (ii) are guaranteed by only assigning feasible subtasks
but not all unassigned subtasks in each loop.
Thirdly, the constraints mentioned
in (iii) and (iv) are considered as a time bound $t_1\in \mathbb{R}^+$
in the bidding process.

(i) \textbf{Initialization}:
Once the R-poset $P_{final}$ is update by \emph{Poset-prod}, we firstly collect
the action sequence $\mathcal{J},\mathcal{J}_o$ of previous solution and
the set of finished subtasks $\Omega_{finish}$ from executing word $w_0$.
Specially, all of them will be empty if it is the first round.
Then, a set of essential conflict subtasks $\Omega_{ec}$ is defined to collect the subtasks which might conflicts the updated partial orders $\preceq_{f}, \neq_{f}$:
\begin{equation}
\begin{aligned}
\Omega_{ec}=&\{\omega_i|  \omega_i\preceq_{f}\omega_j, t_i\geq t_j,\forall\omega_i,\omega_j\in \Omega_1/\Omega_{finish} \}\cup\\
&\{\omega_i|  \omega_i\neq_{f}\omega_j, t_j\leq t_i\leq t_j+d_j,\\&\forall\omega_i,\omega_j\in \Omega_1/\Omega_{finish} \}.
\end{aligned}
\label{get_removeable_set1}
\end{equation}
Then, we compute the set of subtasks $\Omega_{conf}$ that should be removed from $\mathcal{J},\mathcal{J}_o$:
\begin{equation}
\begin{aligned}
\Omega_{conf}=&\{\omega_j|  \omega_i\preceq_{f}\omega_j, \forall\omega_i\in \Omega_1/\Omega_{finish},\\&\forall \omega_j\in \Omega_{ec} \}\cup\Omega_{ec},
\end{aligned}
\label{get_removeable_set2}
\end{equation}
in which are the subtasks in $\Omega_{ec}$ and the subtasks whose pre-subtasks will be removed.
With the action sequences $\mathcal{J},\mathcal{J}_o$ removed all the subtasks in $\Omega_{conf}$, we can initialize the set of assigned subtasks $\Omega_{as}=\{\omega_i|\forall (t_i,\omega_i,a_i) \in \mathcal{J}\}$ and the set of unassigned
subtasks $\Omega_{u}=\Omega_{f}/\Omega_{as}$.

(ii) \textbf{Computation of Feasible Subtasks}:
After initialization, the algorithm starts a loop to assigned the subtasks in $\Omega_{u}$: getting a set of feasible subtasks $\Omega_{fe}$;
calculating their time bounds; choosing the best subtask.
For the ordering constraints $\preceq_c$, if $(\omega_j,\omega_i)\in\preceq_c$,
assigning $(t_{k_j} ,\omega_j ,a_{k_j} )$ to a task sequence $J_i= \cdots(t_{k_i},\omega_i, a_{k_i})$
will violate such constraints. Thus, the $\Omega_{fe}$ based on current $\Omega_{as}$
is defined as:
\begin{equation}
\begin{aligned}
\Omega_{fe}=\{ \omega_i| \omega_i\in \Omega_u,\forall  (\omega_j,\omega_i)\in\preceq_{f}, \omega_j\in\Omega_{as}\},\\
\end{aligned}
\label{Get_feasible_task}
\end{equation}
in which the subtasks may lead to unfeasible action sequences being eliminated.

(iii) \textbf{Online Bidding}:
Then, we will try assigning each subtask in $\Omega_{fe}$ and only choose the one
with the best result.
Without loss of generality, we assume that subtask $\omega_i\in\Omega_{fe}$
requires a label $(C_k)^{u_k}_{k_1,k_2}$, which
means the agents need to execute the behavior $C_k$ from region $W_{k_1}$ to region $W_{k_2}$
using object $u_k$. Any constraint mentioned
in (ii), (iii) and (iv) can be considered as a time bound $t_1\in \mathbb{R}^+$
which means such constraint can be satisfied after $t_1$.
Here, we use three kinds of time bounds:
the global time bound $t^\omega_i$, the object time bound $t^o_{u_k}$
and the set of local time bounds $\boldsymbol{T_s}$.
The global time bound $t^\omega_i$ is the time instance that the
ordering constraints $\preceq_f$ and conflict constraints $\neq_f$ will be satisfied
if behavior $(C_k)^{u_k}_{k_1,k_2}$ is executed after $t^\omega_i$:
\begin{equation}
\begin{aligned} 
t^\omega_i\geq t_j ,& \forall (j,i) \in \preceq_f, \omega_j\in\Omega_{as},\\
t^\omega_i\geq t_\ell+d_\ell ,& \forall \{\omega_i,\omega_\ell,\cdots\}\in\neq_f,\omega_\ell\in\Omega_{as}.
\end{aligned}
\label{global_time_window}
\end{equation}
For the required object $u_k$, assuming its participated last task is $\mathcal{J}^o_{u_k}[-1]=(t_\ell, \omega_\ell)$,
the object time bound $t^o_{u_k}$ should satisfy that:
\begin{equation}
\begin{aligned}
W^p_{u}(t)=W_{k_1}, t \geq t_\ell+d_\ell,, \forall t\geq t^o_{u_k},
\end{aligned}
\label{object_time_window}
\end{equation}
which means the object $u_k$ will be at the starting region $W_{k_1}$ of the current behavior $(C_k)^{u_k}_{k_1,k_2}$ and
ready for it after $t^o_{u_k}$.
Additionally, we set $t^{o}_{u_k}=\infty$ if the object is not at $W_{k_1}$ after
the action sequence $\mathcal{J}^o_{u_k}$, and
we set $t^o_{u_k}=0$ if the behavior $(C_k)^{u_k}_{k_1,k_2}$ does not require object as $u_k=\emptyset$.
The set of local time bound is defined as 
\begin{equation}
\begin{aligned} 
\boldsymbol{T_s}=\{(\mathcal{A}_n, t^a_n)|\mathcal{A}_n=\mathcal{A}_{M_{type}(n)}\cap \mathcal{A}_{C_k}, \\
t^a_n=t_\ell+d_\ell + d_{\mathcal{G}}(k_\ell,k_1),\forall n\in \mathcal{N}\},
\end{aligned}
\label{local_time_window}
\end{equation}
where $\mathcal{A}_n$ is the set of actions that agent $n$ can provide for 
behavior $C_k$, $t^a_n$ is the earliest time agent $n$ can arrive region $W_{k_1}$,
$t_\ell+d_\ell$ is the time when the last subtask $\omega_\ell\in J_n[-1]$ has finished, 
$d_{\mathcal{G}}(k_\ell,k_1)$ is the cost of moving and $W_{k_\ell}$ is the goal region of $\omega_\ell$.
$(\mathcal{A}_n, t^a_n)$ means agent $n$ can begin behavior $(C_k)^{u_k}_{k_1,k_2}$ after time $t^a_n$
by providing one of action $a_\ell\in\mathcal{A}_n$.
Using these time bounds, we can determine the agents and their providing
actions and generate a new party assignment $\mathcal{J}^i,\mathcal{J}^i_o$
to minimize the ending time of subtask $\omega_i$.
The efficiency $\eta$ of each assignment $\mathcal{J}^i,\mathcal{J}^i_o, \omega_i\in\Omega_u$
is calculated, and the subtask with max efficiency will be chosen.
Afterwards, the chosen subtask is removed from $\Omega_{un}$ and added to $\Omega_{as}$.
The action sequences $\mathcal{J},\mathcal{J}_o$ are updated accordingly as
$\mathcal{J}=\mathcal{J}^i, \mathcal{J}_o=\mathcal{J}^i_o$ for the next iteration.

\subsection{Correctness and Completeness Analysis }
\label{complexity_analysis}

\begin{theorem}[Correctness]
	\label{theorem_1}
	Given two R-posets $P_{1}=(\Omega_{1},\preceq_{1},\neq_{1}),
	P_{2}=(\Omega_{2},\preceq_{2},\neq_{2})$
	generated from $\mathcal{B}_{1},\mathcal{B}_{2}$,
	we have $\mathcal{L}(P_j)\subseteq \mathcal{L}(P_{1})\cap \mathcal{L}(P_{2})$, where $P_{j}\in\mathcal{P}_{final}$,
	$\mathcal{P}_{final}=P_1\otimes P_2$. 
\end{theorem}

\begin{proof}
	If a \emph{word} $w=\sigma'_1\sigma'_2\cdots$ satisfies
	$P_j=(\Omega_j,\preceq_j,\neq_j), P_j\in\mathcal{P}_{final}$, it satisfies the
	three conditions mentioned in Def.~\ref{def:language}. In first condition,
	due to the step \textbf{Task Composition} of \emph{Poset-prod},
	we can infer that
	for any $\omega^1_n=(n,\sigma^1_{n},\sigma^{s1}_{n})\in\Omega_{1},$
	there exists $\omega_{n}=(n,\sigma_{n},\sigma^{s}_{n})\in\Omega_{j}$,
	with $\sigma^1_{n}\subseteq \sigma_n,\sigma^{s1}_{n}\subseteq \sigma^s_n$.
	Thus, we have $\sigma^1_{i_1}\subseteq \sigma_{i_1}\subseteq \sigma'_{\ell_1}$ and
	$\sigma^{s1}_{i_1}\cap\sigma'_{m_1}=\emptyset, \forall m_1<\ell_1$,
	which indicates that $w$ satisfies $P_{1}$ for condition 1.
	For the second condition,
	due to ~\eqref{equ:inherit} in step \textbf{Relation Update},
	we have $\preceq_{i}\subseteq \preceq_j$. Thus, we can infer that
	$w$ satisfies $P_{1}$ for the second condition:
	Any $(\omega^1_{i_1},\omega^1_{i_2})\in\preceq_{1},$ we have
	$ (\omega_{i_1},\omega_{i_2})\in\preceq_{j}$, thus $ \exists \ell_1\leq \ell_2,
	\sigma^1_{i_2}\subseteq\sigma_{i_2}\subseteq\sigma'_{\ell_2},$ and
	$\forall m_2< {\ell_2},\sigma^{s1}_{\ell_2}\subseteq\sigma^s_{\ell_2}\subseteq\sigma'_{m_2}$.
	Additionally, for the last condition, as the word $w$ satisfied the $\neq_j$ order of $P_j$.
	We have $\neq_{1}\subseteq\neq_j$ due to~\eqref{neq_calculate}. Thus, the word $w$ also satisfied
	the third condition. In the end, we can conclude that $w$ satisfies $P_{1}$. In the same way, we can
	proof the $w$ also satisfies $P_{2}$. Thus, $\mathcal{L}(P_j)\subseteq \mathcal{L}(P_1)\cap\mathcal{L}(P_2)$
\end{proof}

\begin{theorem}[Completeness]
	Given two R-posets $P_{1},P_{2}$ getting from $\mathcal{B}_{1},\mathcal{B}_{2}$,
	with enough time budget, \emph{Poset-prod} returns
	a set $\mathcal{P}_{final}$ consisting of
	all final product, and its language $\mathcal{L}(\mathcal{P}_{final})=\bigcup_{P_i\in \mathcal{P}_{final}} \mathcal{L}(P_i)$ is equal to
	$\mathcal{L}(\mathcal{P}_{final})=\mathcal{L}(P_{1})\bigcap \mathcal{L}(P_{2})$. 
\end{theorem}
\begin{proof}
	Due to Theorem~\ref{theorem_1}, it holds that
	$\mathcal{L}(\mathcal{P}_{final})\subseteq \mathcal{L}(P_{1})\bigcap \mathcal{L}(P_{2})$.
	Thus, we only need to show that
	$\mathcal{L}(P_{1})\bigcap \mathcal{L}(P_{2})\subseteq \mathcal{L}(\mathcal{P}_{final})$.
	Given a word $w=\sigma'_1\sigma'_2\cdots$ and $w\in \mathcal{L}(P_{1}) \bigcap \mathcal{L}(P_{2})$,
	$w$ satisfies the first condition in Def.~\ref{def:language} for
	both $P_{1}$ and $P_{2}$ that:
	$\forall\omega^1_{i_1}=(i_1,\sigma^1_{i_1},\sigma^{s1}_{i_1})\in\Omega_{1},$ there exists
	$\sigma'_{j_1}$ with $\sigma^1_{i_1}\subseteq\sigma'_{j_1}$ and $ \sigma^{s1}_{j_1}\subseteq\sigma'_{m_1}, \forall m_1<j_1$;
	$\forall\omega^2_{i_2}=(i_2,\sigma^2_{i_2},\sigma^{s2}_{i_2})\in \Omega_{2}$, there exists $\sigma'_{j_2}$
	with $\sigma^2_{i_2}\subseteq\sigma'_{j_2}$ and $ \sigma^{s2}_{j_2}\subseteq\sigma'_{m_2}, \forall m_2<j_2$.
	If $j_1=j_2$, the step in~\eqref{generate_new_subtask2}~of \textbf{Task Composition}
	will generate a subtask $\omega_{i_1}\in\Omega_j$ with $\omega_{i_1}\models\omega^1_{i_1}, \omega^2_{i_2}$,
	and $\sigma_{i_1}\subseteq\sigma'_{j_1},\forall m_3 < j_1, \sigma^s_{j_1}\subseteq\sigma'_{m_3}$.
	If $j_1\neq j_2$, the~\eqref{generate_new_subtask} will generate $\omega_{M_{\Omega}(i_2)}\models\omega^2_{i_2}$,
	with $\sigma_{M_{\Omega}(i_2)}\subseteq\sigma'_{j_2},\forall m_4 < j_2, \sigma^s_{M_{\Omega}(i_2)}\subseteq\sigma'_{m_4}$.
	Thus, there exists $P_j\in\mathcal{P}_{final}$ that satisfies the first condition.
	For the second condition, $\preceq_j$ consisting of two parts generated by ~\eqref{equ:inherit}
	and ~\eqref{neq_update} guarantees that $w\in \mathcal{L}(P_{1})\cap \mathcal{L}(P_{2})$.
	Moreover,~\eqref{new_partial_order} and ~\eqref{omega_task_update} guarantee that
	$w$ does not conflict the self-loop constraints of $P_{1},P_{2}$.
	Thus, the second condition is satisfied.
	Regarding the third condition, since $\neq_j=\neq_1\cap M_\Omega(\neq_2)$ holds in~\eqref{neq_calculate}, $\neq_j$
	is naturally satisfied by~$w$.
	In conclusion, for any $w\in \mathcal{L}(P_{1}) \cap \mathcal{L}(P_{2})$, $w\in \mathcal{L}(P_{final})$ holds
	and vice versa.
	Thus, $\mathcal{L}(\mathcal{P}_{final})$ is equal to $\mathcal{L}(P_{1})\bigcap \mathcal{L}(P_{2})$.
\end{proof}

\section*{Acknowledgments}

\subsection*{Author Contributions}  
Z. Li and M. Guo initiated the idea. Z. Liu designed the algorithms and conducted the experiments. Z. Liu and M. Guo wrote the first draft of the manuscript. Z. Liu, M. Guo, W. Bao and Z. Li commented and revised the manuscript. Z. Li and W. Bao supervised the work. All authors have read, commented, and approved the final manuscript.

\subsection*{Funding} 
This work was supported by the National Key R\&D Program of
China under grants 2022ZD0116401 and 2022ZD0116400, the National Natural Science Foundation of China under grants U2241214, 62373008,
62203017, and T2121002.

\subsection*{Conflicts of Interest}
The authors declare that there is no conflict of interest regarding the publication of this article.

\subsection*{Data Availability}
The authors confirm that the data supporting the findings of this study are available within the article.

\bibliographystyle{IEEEtran}
\bibliography{Poset_prod.bib}
\newpage
\section*{Supplementary Materials}

Movie s1. Simulation and Experiment of dynamic tasks in a hospital environment.

Table s\ref{table:definition_function}. Definitions of agents, behaviors, objects and interested regions.

Table s\ref{table:definition_formulas}. sc-LTL formulas of simulation.

Table s\ref{table:definition_formulas_experiment}. sc-LTL formulas of experiment.

\begin{table*}
	\centering
	\caption{Definitions of agents, behaviors, objects and interested regions}
	\label{table:definition_function}
	\begin{tabular}{| p{2.5cm} | p{8cm}  | p{1cm} |}
		\hline
		\textbf{Agent }& \textbf{Ability}  & \textbf{Label} \\
		\hline
		Junior Doctor & assist operation, transfer, medicine, record, disinfect, clean & $JD$ \\
		\hline
		Senior Doctor	& preside operation, disinfect, medicine  & $SD$ \\
		\hline 
		Nurse& transfer, clean, supply, record& $Nu$\\
		\hline
		\textbf{Interactive Object} & \textbf{Description} & \textbf{Label} \\ \hline
		Family Visitor & A family member visiting the patient & $FV$ \\ \hline
		Vomiting Patient & A patient who suddenly vomited and contaminated the current region  &  $VP$ \\ \hline 
		Senior patient  & A patient with a mild illness   & $SP$ \\ \hline 
		Junior Patient & A patient with a serious illness  & $JP$\\ \hline 
		\textbf{Cooperative Behavior} & \textbf{Ability and Object Requirement} & \textbf{Label}\\\hline 
		primary operate & assist operation: 1,  supply: 2, senior patient: 1 & $P$\\ \hline
		advance operate & assist operation: 2, preside operation: 1, supply: 2, junior patient: 1& $A$ \\ \hline 
		disinfect ground & clean: 1  & $D$ \\ \hline 
		transfer patient & transfer: 2, junior or vomiting patient: 1&  $T$ \\ \hline 
		collect information & record : 1 & $C$ \\ \hline 
		guide visitor &transfer: 1, family visitor or senior patient: 1& $G$\\ \hline 
		supply medicine & medicine: 1, record: 1& $M$ \\ \hline 
		radiate & clean: 1, disinfect: 1& $R$ \\ \hline 
	\end{tabular}
\end{table*}

\begin{table*}
	\centering
	\caption{sc-LTL formulas for simulation }
	\label{table:definition_formulas}
	\begin{tabular}{| p{1.5cm} | p{10cm}|}
		\hline
		\textbf{Offline Formula} & \textbf{Description}\\ \hline 
		$\varphi_{\texttt{b}_1}$ &	$\Diamond(T^{1}_{w1,o1}\land
		\neg P^{1}_{o1,o1} \land \Diamond (P^{1}_{o1,o1} \land \neg R^{\emptyset}_{o1,o1} \land
		\Diamond T^{1}_{o1,w1} \land \Diamond R^{\emptyset}_{o1,o1}))$ :
                                           Eventually transport the patient $1$ from $w1$ to $o1$
                                           and perform a primary operation for him. Then,
		transport him back and radiate to the operation $o1$. \\ \hline 
		$\varphi_{\texttt{b}_2}$ & 
		$\Diamond(C^{\emptyset}_{w3,w3} \land \Diamond (T^{2}_{w3,o4}
		\land \Diamond A^{2}_{o4,o4} \land \neg R^{\emptyset}_{o4,o4}
		\land \Diamond (T^{2}_{o4,w3} \land \neg A^{2}_{o4,o4}\land \Diamond C^{\emptyset}_{w3,w3} )
		\land \Diamond (R^{\emptyset}_{o4,o4}\land \Diamond D^{\emptyset}_{o4,o4})))$ :
		Record the situation of inpatient ward $w3$ first. Then transport patient $2$ from $w3$ to operation $o4$
		and perform an advanced operation for this patient at $o4$. After that, transport the patient back to $w3$ and Record the situation;
		radiate and disinfect the ground of $o4$. 
		\\\hline  
		$\varphi_{\texttt{b}_3}$ & $\Diamond( C^\emptyset_{w3,w3}\land\neg T^3_{w3,e2}\land\Diamond D^\emptyset_{w3,w3}\land\Diamond T^3_{w3,e2})$ :
		Record the situation of ward $w3$ and transport the patient $3$ into the exit $e2$ and disinfect the ground of $w3$. \\ \hline 
		$\varphi_{\texttt{b}_4}$ & $\Diamond D^{\emptyset}_{w7,w7}  \land \Diamond(C^\emptyset_{w7,w7}
		\land \neg M^{\emptyset}_{w7,w7}
		\land \Diamond M^{\emptyset}_{w7,w7} ):$ 
		Disinfect the ground of ward $w7$, record the information and supply the medicines at $w7$.
		\\ \hline   
		$\varphi_{\texttt{b}_5}$ & $\Diamond (C^\emptyset_{w7,w7} \land \neg G^4_{w7,e3} \land
		\Diamond G^4_{w7, e3}) \land \neg D^\emptyset_{w7,w7} \textsf{U} C^\emptyset_{w7,w7}: $  
		Collect information in ward $w7$ then guide the patient $4$ from ward $w7$ to
		exit $e3$. Do not disinfect the ground of $w7$ before collecting the information.
		\\ \hline
		$\varphi_{\texttt{b}_6}$ & $\Diamond (T^{5}_{h5,w5} \land \Diamond C^{\emptyset}_{w5,w5} \land \Diamond
		M^{\emptyset}_{w5,w5})\land\neg M^{\emptyset}_{w5,w5} \textsf{U} C^{\emptyset}_{w5,w5} $:
		Transport patient $5$ form hall $h5$ to $w5$ and supply medicine and collect information of $w5$.
		Do not supply medicine before collect the information.\\ \hline 
		\textbf{Online Formula}& \textbf{Description}\\ \hline
		$\varphi_{\texttt{VP}}$&
		$\Diamond (T^u_{i,i} \land \Diamond R^\emptyset_{i,i} \land \Diamond D^\emptyset_{i,j}) \land
		\neg Nu_{i}  \textsf{U} R^\emptyset_{i,i} $: When a patient has vomited in region $i$, transport him
		to region $j$, radiate and disinfect region $i$. Nurses should not enter region $i$ before radiate.\\ \hline  
		$\varphi_{\texttt{JP}}$& $ \Diamond G^u_{i,j} \land \Diamond R^\emptyset_{i,i} \land  \neg
		JD_{i} \textsf{U} R^\emptyset_{i,i} :$ When a junior patient comes to region $i$, guide him to
		region $j$ and radiate the region $i$, the junior doctor should not enter $i$ until radiate.  \\ \hline
		$\varphi_{\texttt{SP}}$ & $\Diamond (T^u_{i,j} \land (\bigwedge_{\ell=1\cdots 3} \neg SP_{o_\ell})
		\land \Diamond M^\emptyset_{j,j} \land \Diamond C^\emptyset_{j,j}) :$ When a senior patient comes to region $i$, transport him to region $j$ and do not
		enter operation rooms $o1,o2,o3$; then supply medicine and collect information. 
		\\ \hline
		$\varphi_{\texttt{FV}}$& $ \Diamond G^u_{i,j} \land( \bigwedge_{\ell=1\cdots5}\neg  FV_{o_\ell}) :$ When a family visitor comes,
		guide him to his goal and do not enter the operation rooms $o_1,\cdots,o_5$.\\ \hline
	\end{tabular}
\end{table*}

\begin{table*}
	\centering
	\caption{sc-LTL formulas for experiment}
	\label{table:definition_formulas_experiment}
	\begin{tabular}{| p{1.5cm} | p{10cm}|}
		\hline
		\textbf{Formula} & \textbf{Description}\\ \hline 
		$\varphi_{\texttt{b}_1}$  &
		$ \Diamond (C^{\emptyset}_{w3,w3}  \land \Diamond(T^1_{w3,o4}
		\land \Diamond ( A^{1}_{o4,o4} \land\neg R^{1}_{o4,o4} \land \Diamond (T^1_{o4,w3} \land
		\neg A^1_{o4,o4} \land\Diamond C^{\emptyset}_{w3,w3})) \land \Diamond R^{\emptyset}_{o4,o4})) :$ 
                                            Collect information in $w3$, transport the patient $1$ from $w3$ to $o4$
                                            and perform an advanced operation for him. Then,
		transport him back and collect the information; radiate $o4$.\\ \hline 
		$\varphi_{\texttt{b}_2}$&$\Diamond(C^{\emptyset}_{w3,w3}\land
		\neg T^{2}_{w3,e2} \land \Diamond D^{\emptyset}_{w3,w3} \land \Diamond T^{2}_{w3,e2}) :$
		Collect information in $w3$, transport the patient $2$ from $w3$ to $e2$ and disinfect the area $w3$.\\\hline 
		$ \varphi_{\texttt{b}_3}$ & $\Diamond (T^3_{h5,w2} \land \Diamond C^\emptyset_{w2,w2}\land \Diamond M^\emptyset_{w2,w2}) :$
		Transport patient $3$ from $h5$ to $w2$ then collect the information in $w2$, and supply medicine at $w2$.
		\\ \hline 
		$\varphi_{\texttt{b}_4}$ & $\neg M^\emptyset_{w2,w2} \textsf{U} C^\emptyset_{w2,w2} :$ Do not supply medicine at $w2$ before
		collecting information. \\ \hline
		$\varphi_{\texttt{VP}}$&
		$\Diamond (T^u_{i,i} \land \Diamond R^\emptyset_{j,j} \land \Diamond D^\emptyset_{i,j}) \land
		\neg Nu_{i}  \textsf{U} R^\emptyset_{i,i} $: When a patient has vomited in region $i$, transport him
		to region $j$, radiate and disinfect region $i$. Nurses should not enter region $i$ beforehand.\\ \hline  
		$\varphi_{\texttt{JP}}$& $ \Diamond G^u_{i,j} \land \Diamond R^\emptyset_{i,i} \land  \neg
		JD_{i} \textsf{U} R^\emptyset_{i,i} : $ When a junior patient comes to region $i$, guide him to
		region $j$ and radiate the region $i$, the junior doctor should not enter $i$ beforehand.  \\ \hline
		$\varphi_{\texttt{SP}}$ & $\Diamond (T^u_{i,j} \land (\bigwedge_{\ell=1\cdots 3} \neg SP_{o_\ell})\land \Diamond M^\emptyset_{j,j} \land \Diamond C^\emptyset_{j,j}) :$ When a senior patient comes to region $i$, transport him into region $j$ and do not
		enter operation rooms $o1,o2,o3$; then supply medicine and collect information. 
		\\ \hline
		$\varphi_{\texttt{FV}}$& $ \Diamond G^u_{i,j} \land( \bigwedge_{\ell=1\cdots5}\neg  FV_{o_\ell}) :$ When a family visitor comes,
		guide him to his goal and do not enter the operation rooms $o_1,\cdots,o_5$.\\ \hline
	\end{tabular}
\end{table*}


\end{document}